\documentclass[journal,compsoc]{IEEEtran}

\ifCLASSOPTIONcompsoc
   \usepackage[nocompress]{cite}
\else
   \usepackage{cite}
\fi
\usepackage{graphicx}
\usepackage{amsmath,amssymb,amsthm}
\usepackage{stmaryrd}

\usepackage[export]{adjustbox}
\newcommand\independent{\protect\mathpalette{\protect\independenT}{\perp}}
\def\independenT#1#2{\mathrel{\rlap{$#1#2$}\mkern2mu{#1#2}}}

\newcommand{\norm}[1]{\left\lVert#1\right\rVert}
\newcommand{\R}{\mathbb{R}}

\usepackage{subfig}
\usepackage{algorithm2e}
\SetKw{KwBy}{by}

\newtheorem{theorem}{Theorem}[section]

\newtheorem{lemma}[theorem]{Lemma}
\DeclareMathOperator{\Tr}

\usepackage{pgfplots}
\usepackage{tikz}
\usetikzlibrary{3d}
\usetikzlibrary{positioning}
\usetikzlibrary{calc,matrix,backgrounds,fit}
\pgfdeclarelayer{myback}
\pgfsetlayers{myback,background,main}

\usepackage{comment}

\begin{document}
\title{Column-wise Element Selection for Computationally Efficient Nonnegative Coupled Matrix Tensor Factorization}
%\author{Author Name}
\author{Thirunavukarasu~Balasubramaniam,%~\IEEEmembership{Member,~IEEE,}
       ~Richi~Nayak,%~\IEEEmembership{Member,~IEEE,}
        ~Chau~Yuen,%~\IEEEmembership{Member,~IEEE,}
        ~and~Yu-Chu~Tian%,~\IEEEmembership{Member,~IEEE}% <-this % stops a space
\IEEEcompsocitemizethanks{\IEEEcompsocthanksitem T. Balasubramaniam, R. Nayak, and Y.-C. Tian are with the School of Electrical Engineering and Computer Science, Queensland University of Technology, Australia,
QLD, 4000.\protect\\
E-mail: {thirunavukarasu.balas, r.nayak, y.tian}@qut.edu.au 
\IEEEcompsocthanksitem C.Yuen is with Singapore University of Technology and Design, Singapore. Email: yuenchau@sutd.edu.sg%.\protect\\
%Email: yuenchau@sutd.edu.sg
}% <-this % stops an unwanted space
}
\markboth{IEEE Transactions on Knowledge and Data Engineering}{Balasubramaniam \MakeLowercase{\textit{et al.}}: Column-wise Element Selection for Computationally Efficient Nonnegative Coupled Matrix Tensor Factorization}

\IEEEtitleabstractindextext{%
\begin{abstract}
Coupled Matrix Tensor Factorization (CMTF)  facilitates the integration and analysis of multiple data sources and helps discover meaningful information.  Nonnegative CMTF (N-CMTF) has been employed in many applications  for identifying latent patterns, prediction, and recommendation. However, due to the added complexity with coupling between tensor and matrix data, existing N-CMTF algorithms exhibit poor computation efficiency. In this paper,  a computationally efficient N-CMTF factorization algorithm is presented based on the column-wise element selection, preventing frequent gradient updates. %However, a direct use of N-CMTF for pattern mining applications with sparse data may generate less accurate or meaningless patterns. To address this issue, a sparsity constrained Context-Aware Nonnegative Coupled Sparse Matrix Tensor Framework (CAN-CSMTF) is further developed. It is able to identify true factors in the sparse dataset. 
Theoretical and empirical analyses show that the proposed N-CMTF factorization algorithm is not only more accurate but also more computationally efficient than existing algorithms in approximating the tensor as well as in identifying the underlying nature of factors.
\end{abstract}

}

\maketitle
 
\IEEEdisplaynontitleabstractindextext

\IEEEpeerreviewmaketitle

%===============
\IEEEraisesectionheading{\section{Introduction}\label{sec:introduction}}

\IEEEPARstart {W}{ith} the digital advancements, %have led to the generation of astronomical amounts of data. 
the data about the same concept can be collected from multiple sources. %Multidimensional tensor models have been successfully used to capture multi-type relationships exhibited in real-world datasets \cite{symeonidis2016matrix}. 
To get meaningful information on a topic, it is important to combine and analyze relevant data accurately and efficiently from multiple sources. This is known as data fusion, which is a fundamental technique widely used in a wide range of domains such as recommender systems \cite{acar2011all}, pattern mining \cite{balasubramaniam2018nonnegative}, metabolomics \cite{acar2015data}, chemometrics \cite{acar2013understanding}, sensor and signal processing \cite{hunyadi2014block}, brain imaging \cite{cichocki2009nonnegative}, and bioinformatics \cite{iakovidou2010multiway}. 

In recommender systems, the “user-to-item rating” ($user \times item$) information can be fused with the “user-to-user trust” ($user \times user$) information. %This coupled matrix fusion leads to an improved accuracy of rating estimation~\cite{acar2011all}. 
Similarly, the multifaceted nature of datasets facilitates the fusion of matrix and tensor data sources for discovering useful knowledge.  To recommend an item to a user based on its tagging activities, the primary data source can be represented as a third-order tensor ($user \times item \times tag$) with an auxiliary trust matrix ($user \times user$). In smart city applications, people's check-in activities can be represented as a third order tensor ($user \times location  \times time$) with a ($location \times category$) matrix. The coupled matrix-tensor fusion models are shown to be useful in real applications~\cite{balasubramaniam2018nonnegative,acar2011all,acar2013understanding}. 

Traditional Matrix Factorization (MF) only or Tensor Factorization (TF) only algorithms such as Alternating Least Square (ALS) \cite{cichocki2009nonnegative}, Stochastic Gradient Descent (SGD) \cite{zinkevich2010parallelized}, Multiplicative Update Rule (MU) \cite{lee2001algorithms}, and Coordinate Descent (CD) \cite{wright2015coordinate} can fail to capture the latent factors with interrelated multi-dimensions data like coupled matrix-tensor \cite{cai2011graph}. A coupled matrix-tensor factorization (CMTF)~\cite{acar2011all} and a Nonnegative CMTF (N-CMTF)~\cite{turbo} have been specifically proposed to jointly analyze matrix and tensors. However, their computation efficiency and convergence speed are not acceptable particularly for sparse data~\cite{bader2007efficient}. While these algorithms can be implemented in parallel and distributed environments to improve the scalability, same as TF \cite{turbo,flexifact,papalexakis2013scoup,oh2017s,shin2017fully}, the underlying computational complexity remains the same. With the rise in this type of data, there exists a need to develop a computationally efficient N-CMTF algorithm.

In this paper, we propose a column-wise element selection-based CD algorithm, Cut-off Coordinate Descent (Cut-CD), to solve N-CMTF efficiently and accurately. Cut-CD updates factor matrices one by one. While updating a factor matrix, it selects only a few important elements based on a column-wise cut-off technique, instead of updating all the elements. It calculates each element’s importance based on the difference in the objective function that measures the error minimized using optimization. %For each factor matrix, it derives a single element update rule for N-CMTF. 
By using the column-wise element selection technique, Cut-CD avoids the frequent gradient updates~\cite{gcd2011}, a bottleneck in traditional element selection-based methods, and speeds up the factor matrix update process. For the purpose of pattern mining, we propose to introduce sparsity constraint to the N-CMTF objective function in Cut-CD (called as Cut-CD-SC) to capture accurate factor matrices. With an efficient N-CMTF algorithm as Cut-CD, the shared and unshared latent factors can be learned accurately, which can facilitate accurate prediction, recommendation and pattern mining for large datasets.%For the purpose of pattern mining, we propose to introduce sparsity constraint to N-CMTF model in Cut-CD (called as Cut-CD-SC) to capture accurate factor matrices. With an efficient N-CMTF algorithm as Cut-CD, the shared and unshared latent factors can be learned accurately that can facilitate accurate prediction, recommendation and pattern mining that can be applied to large datasets. 

Extensive empirical analysis is conducted to show the efficiency of Cut-CD for the tasks of tensor approximation and recommendation, and the accuracy of Cut-CD-SC for spatio-temporal pattern mining in sparse datasets. Using synthetic and real-world datasets, we demonstrate 1) the fast convergence speed of Cut-CD; 2) the efficiency of Cut-CD as compared to other state-of-the-art algorithms for accurately generating recommendations and finding the underlying structure of spatio-temporal patterns; and (3) the accuracy of Cut-CD-SC in finding distinctive spatio-temporal patterns.

In summary, this paper presents the followings:
     \begin{enumerate}\item[1)]\textbf{Factorization Algorithm.} Column-wise element selection-based Cut-CD algorithm for N-CMTF. It updates the elements and gradients selectively, leading to faster convergence without a trade-off of accuracy.
     
    %\item[2)] \textbf{Objective Function.} A Sparsity Constraint is incorporated into the N-CMTF model. As the input data is usually sparse, our proposed CAN-CSMTF introduces sparsity constraint to the N-CMTF process to derive sparse factor matrices. In addition to sparsity constraint, multiple contexts available in the data are also incorporated as N-CMTF that makes CAN-CSMTF a context-aware model. This improves the quality of the factor matrices revealing spatio-temporal patterns. 
     
     \item[2)]\textbf{Theoretical analysis.} Theoretical results are established for the efficiency of Cut-CD in terms of convergence, time complexity, memory requirement, and element/gradient updates.
     
     \item[3)]\textbf{Experiments.} Extensive empirical analysis is conducted to show the efficiency of Cut-CD in terms of runtime and convergence speed as well as accuracy in the tasks of approximation, prediction and pattern mining.
 \end{enumerate}

The rest of the paper is organized as follows: Section 2 reviews related work. Section 3 introduces preliminaries for TF and CMTF. Section 4 presents Cut-CD. Theoretical analysis and empirical analysis are discussed in Sections 5 and 6 respectively. %Finally, Section 7 concludes the paper.

\section{Related Work}
\textbf{CMTF Applications:} With the wide availability of multiple data sources (e.g., mobile computing and Internet-of-Things devices), CMTF has become more effective in missing data estimation (i.e. prediction) or pattern mining in comparison to using MF or TF independently \cite{bhargava2015and,zheng2014diagnosing}. Tensor modeling ($location \times noise~category \times time~slot$) was applied to understand the noise pollution across New York City to inform people about the better environment \cite{zheng2014diagnosing}. The results show that the traditional TF fails to predict most of the missing entries due to the data being sparse. On the other hand, a CMTF method was successfully used to recommend an activity, time and location to a user by modelling ($user \times activity \times location \times time$) with additional information coupled as a matrix at each mode \cite{bhargava2015and}. CMTF has also been used in location recommendation by combining a ($user \times location \times time$) tensor model with a ($user \times user$) similarity matrix and a ($location \times feature$) matrix \cite{zhu2016context}. The runtime of the method was improved by utilizing a threshold algorithm that separates the data into smaller regions and calculates the top-recommendation for each region. Recently, N-CMTF was applied to predict spatio-temporal patterns using Greedy Coordinate Descent (GCD) \cite{balasubramaniam2018nonnegative}. The method highlighted the scalability issues with N-CMTF particularly in the presence of high sparse data. In general, all of these methods have been applied to small datasets only. In reality, the majority of tensor applications results in a large size and very high sparsity dataset \cite{bhargava2015and,acar2012metabolomics}.

\textbf{N-CMTF algorithms:} %Nonnegative tensor or matrix factorization (NTF/NMF) algorithms have become popular due to their capacity to accurately analyse the hidden latent relationships in the data represented as tensor or matrix form \cite{luo2016nonnegative}. 
Traditionally, ALS has been used to solve Nonnegative Tensor or Matrix Factorization (NTF/NMF) problems. Because of the non-convex optimization problem underlying the objective function, only one factor matrix is updated by fixing all the other factor matrices in ALS to solve the optimization problem. This approach of ALS has been extended to solve the N-CMTF because of its simplicity in formulation \cite{acar2011all}. However, due to the expensive matrix multiplications involving in the update rule, ALS becomes non-scalable and prone to poor convergence~\cite{comon2009tensor,kolda2009tensor}. To overcome poor convergence, all-at-once optimization (OPT)~\cite{acar2011all} was introduced for CMTF that updates all factor matrices together. This gradient-descent based approach has been reported unable to identify the true underlying factors of the factor matrices \cite{acar2011all,acar2013understanding}. 
% process enforcing the many matrix and tensor factorization methods have been proposed. These methods either modify the objective function or factorization algorithm depending on the application. For example, the authors in  modified the NMF objective function by incorporating a graph regularization term for clustering problem. The factorization algorithm is then modified to effectively learn the graph structure that improves the clustering accuracy. 
%Similarly, the authors in \cite{imtiaz2018distributed} proposed a privacy preserving factorization algorithm to solve NMF and NTF in distributed systems.

\textbf{Fast NMF/NTF algorithms:} The time complexity involved in the traditional factorization process is mainly because of the Matricized Tensor Times Khatri-Rao Product (MTTKRP) and the factor matrix update~\cite{TTB_Sparse}. 

To effectively parallelize or distribute the calculation of MTTKRP, the majority of researchers rely on the parallel and distributed setups \cite{turbo,flexifact,papalexakis2013scoup,oh2017s,shin2017fully,imtiaz2018distributed}. For instance, the parallel and distributed stochastic gradient-based FlexiFact algorithm was proposed to reduce the computational cost \cite{flexifact}. While these approaches may improve the performance in a distributed environment, they hardly have an improvement on traditional non-distributed machines. The random sampling based method called Turbo-SMT was used to reduce the size of the input tensor \cite{turbo}. The sampled tensor minimizes the size of the processing tensor involving MTTKRP, thus the computation cost is reduced. However, the random sampling of an original tensor to a smaller subset is prone to information loss and may result in factors learned to be inaccurate. %is prone to information loss, thus it results in factors learned to be inaccurate.

On the other hand, the \textbf{Coordinate Descent (CD)} based NMF/NTF algorithms have been developed to reducing the time complexity by involving the efficient factor matrix updates \cite{cicho2009fhals,richtarik2018parallel}. They have shown better accuracy and convergence property than ALS and OPT \cite{kim2014alg,kimura2015var}. %The CD-based algorithms are also known easy to be parallelized \cite{richtarik2018parallel}. 
The optimization is solved as a single element sub-problem and each factor matrix is alternatively element-wise updated \cite{cicho2009fhals}. This minimizes the computational cost of factor matrix updates. This element-wise update leads the algorithm to choose and update important elements repeatedly that proves to converge faster. Hsieh et.al. introduced GCD algorithm to solve the NMF problem \cite{gcd2011} where the elements to be updated are greedily selected based on the importance measure calculated using the gradients. GCD updates a single element multiple times and, for each element update, it recalculates the gradient value of the entire row. However, this gradient recalculation significantly increases the computational cost and memory requirements. In comparison, the proposed Cut-CD selects a set of elements to be updated in each column and avoids frequent gradient updates that are a bottleneck in GCD. We have implemented GCD which is the only element selection-based algorithm in N-CMTF and used it as a benchmark in experiments.

%Coordinate Descent for Tensor Factorization (CDTF)~\cite{shin2017fully,shin2014distributed} and Parallel Collective Matrix Factorization (PCMF)~\cite{rossi2016parallel} are two closely related work that is based on (Cyclic Coordinate Descent++) CCD++~\cite{ccdpp}. Without considering the parallelization and distribution, both CDTF and PCMF are same except the fact that CDTF is the extension of CCD++ for higher orders whereas PCMF is the extension of CCD++ for collective factorization. The difference between the proposed Cut-CD and CCD++ based Matrix Tensor factorization methods like CDTF and PCMF are three-fold.
CCD++~\cite{ccdpp} is a CD-based MF algorithm that updates the $k^{th}$ column of each factor matrix without any element selection and repeats the $k^{th}$ column updating for other factor matrices.  This process is then done for all columns. This is considered as a single iteration, where the method requires multiple such iterations. Subset Alternating Least Square (SALS) ~\cite{ccdpp} is an intermediate version of ALS and CCD++ with an additional constraint that controls the number of columns to update in a single iteration of CCD++. However, SALS cannot be directly applied for N-CMTF due to the additional constraint it imposes. Coordinate Descent for Tensor Factorization (CDTF)~\cite{shin2017fully,shin2014distributed} and Parallel Collective Matrix Factorization (PCMF)~\cite{rossi2016parallel} are two closely related work in which the underlying factorization algorithm is CCD++~\cite{ccdpp}. Without considering the parallelization and distribution, both CDTF and PCMF are the same except the fact that CDTF is the extension of CCD++ for higher orders whereas PCMF is the extension of CCD++ for collective factorization. We have implemented CCD++ in N-CMTF environment and used it as a benchmark in experiments. The difference between the proposed Cut-CD and CCD++ based matrix/tensor factorization algorithms are three-fold.
\begin{enumerate}
    \item[1)] Cut-CD first selects elements in a factor matrix according to element importance based on the proposed column-wise cut-off technique. All the selected elements of the factor matrix are then updated. This process is then repeated for the rest of the matrices sequentially. This forms an iteration.  Whereas CCD++ and its dependent methods update the $k^{th}$ column of each factor matrix without any element selection and repeats the $k^{th}$ column updating for other factor matrices.  This process is then done for all columns. This forms a single iteration. This frequent change in factor matrix is repeated for $R$ (number of columns) times which involves higher communication cost making its convergence slower.  In summary, Cut-CD and CCD++ considerably vary in terms of how column elements are selected and how the sequence of columns are updated. 
	
	%\item[1)] The update sequence of CutCD is similar to ALS where all the entries of a factor matrix is updated and then move on to the next factor matrix. In CCD++, the entries in, for example, the kth column of a factor matrix is updated and then move on to the kth column of the next factor matrix. After the kth column of all the factor matrix are updated, the k+1th column of the matrices are updated. This frequent change in the factor matrix involves higher communication cost making it converge slower.
	\item[2)]CCD++ involves multiple inner iterations during the column-wise update whereas Cut-CD does not have any inner iteration. These inner iterations cause CCD++ to update the same column multiple times which introduces frequent gradient updates, a time consuming task similar to GCD.
	%\item[3)]CCD++ requires an accelerating technique to control the inner iterations whereas Cut-CD does not require that. The column-wise element selection makes Cut-CD to converge faster without any inner iterations.
	\item[3)]CCD++ relies on an accelerating technique (stopping criteria) to control the inner iterations and to avoid unnecessary repetitive update of elements. On the other hand, the column-wise element selection makes Cut-CD converge faster without any inner iterations or repetitive update of elements.
\end{enumerate}

In summary, though many factorization algorithms have been proposed for matrix/tensor factorization, an efficient algorithm for N-CMTF is yet to be seen. The existing NMF/NTF algorithms fail to address the computational cost associated with the factor matrix update. The NMF/NTF algorithms compromise either with accuracy or need expensive distributed setup while providing an efficient solution. Recently a family of methods has focused on combining manifold learning with NMF/NTF to improve the accuracy of the factorization process by incorporating the graph regularization term to capture the closeness information in the data~\cite{chen2018adaptive,luong2018novel,luong2019clustering}. These methods alter the objective function according to domain-specific requirements, but the underlying factorization algorithm remains similar to ALS. The process of building a nearest graph is very expensive, thus these algorithms are limited to small data size. The purpose of Cut-CD is to improve the underlying factorization process as used in all matrix and tensor factorization. Cut-CD is designed for Euclidean distance objective function that is commonly used. Therefore, the overarching aim is to utilise Cut-CD in the factorization process of these methods to achieve a computationally efficient performance. Cut-CD can be easily adapted for NMF or NTF problems in general.

\section{Coupled Matrix Tensor Factorization}
The notations used throughout the paper are summarized in Table \ref{tab:notations}. Let \(\boldsymbol{\mathcal{X}} \in \R^{(J\times K\times L)} \) denote a third-order tensor where \(J\), \(K\) and \(L\) represent the length of the dimensions (or modes) of the tensor.

\begin{table}[tb!]
\caption{Notations used in thie paper.}
\label{tab:notations}
\centering
\begin{tabular}{cl}
\hline
Notation & Description\\
\hline
\( \boldsymbol{\mathcal{X}} \)   & tensor (Euler script letter)\\
\( \Omega \)   & set of indices of observable entries of \( \boldsymbol{\mathcal{X}} \)\\
\( x_{jkl} \)   & $(j,k,l)^{th}$ entry of \( \boldsymbol{\mathcal{X}} \)\\
\( \boldsymbol{\mathrm{U}}\) & matrix (upper case, bold letter)\\
\( \boldsymbol{\mathrm{u}}\) & vector (lower case, bold letter)\\
\( \mathit{u}\) & scalar (lower case, italic letter) / element\\
\( \boldsymbol{\mathrm{X_n}}\) & mode-n matricization of tensor\\
\(\otimes\) & Kronecher product\\
\(\odot\) & Khatri-Rao product\\
\(\ast\) & Hadamard product\\
\(\circ\) & outer product\\
\(\norm{.}\) & Frobenius norm\\
\hline
\end{tabular}
\end{table}

%\textbf{Tensor.} Tensor is an extension of matrix to represent higher-order data that describes the elements of the multi-linear space. Vector is a tensor of order 1 and matrix is a tensor of order 2. The structure greater than 2 is called higher-order tensor. 

\textbf{Matricization of the tensor.} The process of converting a tensor into a matrix is called as matricization or unfolding of tensor \cite{tucker1966}. The mode-1 matricization can be denoted as \( \boldsymbol{\mathrm{X_1}}\ \in \R^{(J \times (KL))} \) for a third order tensor.

%Several matrix products are required for the tensor-based processes \cite{delatauwer}. We briefly introduce the main ones. 

\textbf{Kronecker product.} For two matrices denoted as \( \boldsymbol{\mathrm{U}} \in \R^{(J \times R)}\)  and \( \boldsymbol{\mathrm{V}} \in \R^{(K \times R)}\), the Kronecker product is presented as \(\boldsymbol{\mathrm{U}} \otimes \boldsymbol{\mathrm{V}} \). The resultant matrix of size \((JK \times R^2)\)  is defined as follows:
\begin{equation}
\label{eq_1}
\boldsymbol{\mathrm{U}} \otimes  \boldsymbol{\mathrm{V}} = \begin{bmatrix}
    u_{11}\boldsymbol{\mathrm{V}} & u_{12}\boldsymbol{\mathrm{V}} & u_{13}\boldsymbol{\mathrm{V}} & \dots  & u_{1r}\boldsymbol{\mathrm{V}} \\
    u_{21}\boldsymbol{\mathrm{V}} & u_{22}\boldsymbol{\mathrm{V}} & u_{23}\boldsymbol{\mathrm{V}} & \dots  & u_{2r}\boldsymbol{\mathrm{V}} \\
    \vdots & \vdots & \vdots & \ddots & \vdots \\
    u_{j1}\boldsymbol{\mathrm{V}} & u_{j2}\boldsymbol{\mathrm{V}} & u_{j3}\boldsymbol{\mathrm{V}} & \dots  & u_{jr}\boldsymbol{\mathrm{V}}
\end{bmatrix}
\end{equation}
\begin{equation}
  \label{eq_2} 
  = \begin{bmatrix} 
  \boldsymbol{\mathrm{u_{1}}}\boldsymbol{\mathrm{V}} & \boldsymbol{\mathrm{u_{2}}}\boldsymbol{\mathrm{V}} & \boldsymbol{\mathrm{u_{3}}}\boldsymbol{\mathrm{V}} & \dots  & \boldsymbol{\mathrm{u_{r}}}\boldsymbol{\mathrm{V}} \\
  \end{bmatrix}  
\end{equation}
where \( \boldsymbol{\mathrm{u_r}} \) and \( \boldsymbol{\mathrm{v_r}} \) are the columns of the matrices \(\boldsymbol{\mathrm{U}}\) and \(\boldsymbol{\mathrm{V}}\) respectively.

\textbf{Khatri-Rao Product.} Column-wise Kronecker product is called as Khatri-Rao product and it is denoted as \(\boldsymbol{\mathrm{U}} \odot \boldsymbol{\mathrm{V}}\). The resultant matrix is of size \( (JK \times R) \) and is defined as: 
\begin{equation}
    \label{eq_3}
    \boldsymbol{\mathrm{U}} \odot  \boldsymbol{\mathrm{V}} = 
    \begin{bmatrix}
        \boldsymbol{\mathrm{u_1}} \otimes \boldsymbol{\mathrm{v_1}} & \dots & \boldsymbol{\mathrm{u_r}} \otimes \boldsymbol{\mathrm{v_r}}.
    \end{bmatrix}
\end{equation}

\textbf{Hadamard Product.} When the size of two matrices are the same, the Hadamard product can be calculated by the element-wise matrix product. It is denoted by \(\boldsymbol{\mathrm{U}} \ast  \boldsymbol{\mathrm{V}}\) and defined as:
\begin{equation}
%\label{eq_4}
 \boldsymbol{\mathrm{U}} \ast  \boldsymbol{\mathrm{V}} = \begin{bmatrix}
    u_{11} v_{11} & u_{12} v_{12} & u_{13} v_{13} & \dots  & u_{1r} v_{1r} \\
    u_{21} v_{21} & u_{22} v_{22} & u_{23} v_{23} & \dots  & u_{2r} v_{2r} \\
    \vdots & \vdots & \vdots & \ddots & \vdots \\
     u_{j1} v_{j1} & u_{j2} v_{j2} & u_{j3} v_{j3} & \dots  & u_{jr} v_{jr}. 
\end{bmatrix}
\end{equation}

\textbf{Tensor Factorization} is a dimensionality reduction technique that factorizes the given tensor into factor matrices that contain latent features. CANDECOMP/PARAFAC (CP) factorization and Tucker factorization are two well-known tensor factorization techniques \cite{kolda2009tensor}. CP factorization has shown to be less expensive in both memory and time consumption as compared to Tucker factorization \cite{kolda2009tensor}.  Therefore, CP factorization has been more commonly used.

\textbf{\textit{Definition 1} (CP Factorization): } For a tensor \(\boldsymbol{\mathcal{X}} \in \R^{(J\times K\times L)} \) and rank \(R\), the CP factorization factorizes the tensor into a sum of component rank-one tensors~cite{carroll1970} as:
\begin{equation}
\label{eq_5}
    \boldsymbol{\mathcal{X}} \cong \llbracket \boldsymbol{\mathrm{U}}, \boldsymbol{\mathrm{V}}, \boldsymbol{\mathrm{W}} \rrbracket = \sum_{r=1}^{R} \boldsymbol{\mathrm{u_{r}}} \circ \boldsymbol{\mathrm{v_{r}}} \circ \boldsymbol{\mathrm{w_{r}}}.
\end{equation}

\begin{figure}[!t]
\centering
\begin{tikzpicture}
        \tikzset{xzplane/.style={canvas is xz plane at y=#1, thick,shading=axis,shading angle=90,top color = blue!30,draw=black}};
        \tikzset{yzplane/.style={canvas is yz plane at x=#1,thick,shading=axis,shading angle=90,top color = green!30,draw=black}};
        \tikzset{xyplane/.style={canvas is xy plane at z=#1,thick,shading=axis,shading angle=90,top color = red!50,draw=black}};
        \tikzset{xyplane2/.style={canvas is xy plane at z=#1,thick,shading=axis,shading angle=90,top color = red!50,draw=black}};
          
        \draw[xzplane=1.2] (0,0)--(1.2,0)--(1.2,1.2)--(0,1.2) --cycle;
        \draw[yzplane=1.2] (0,0)--(1.2,0)--(1.2,1.2)--(0,1.2) --cycle;
        \draw[xyplane=1.2] (0,0)--(1.2,0)--(1.2,1.2)--(0,1.2) --cycle;      
        \draw[xyplane=2] node[black] at (1,1){\(\boldsymbol{\mathcal{X}}\)};
        \begin{scope}[shift={(1.9 cm,0 cm)}]
            \draw[xyplane=2.2] node[black] at (0.5,1.5){\Large\(\boldsymbol{\cong}\)};
        \end{scope}
        
        \begin{scope}[shift={(2.3 cm,0 cm)}]
            \draw[xyplane=1] (0,0)--(0.15,0)--(0.15,1)--(0,1) --cycle;
            \draw[xyplane=1, top color = blue!30] (0.21,1)--(1.21,1)--(1.31,1.15)--(0.31,1.15) --cycle;
            \draw[xyplane=1, top color = green!30] (0.1,1.16)--(0.25,1.16)--(0.80,1.8)--(0.65,1.8) --cycle;
            \draw[xyplane=1] node[black] at (0.17,-0.2){\(\boldsymbol{\mathrm{u_1}}\)};
            \draw[xyplane=1] node[black] at (1.3,1.34){\(\boldsymbol{\mathrm{v_1}}\)};
            \draw[xyplane=1] node[black] at (0.8,2){\(\boldsymbol{\mathrm{w_1}}\)};
        \end{scope}
        
        \begin{scope}[shift={(3.8 cm,0 cm)}]
            \draw[xyplane=2] node[black] at (0.5,1.5){\Large\(+\)};
        \end{scope}
        \begin{scope}[shift={(4.2 cm,0 cm)}]
            \draw[xyplane=1] (0,0)--(0.15,0)--(0.15,1)--(0,1) --cycle;
            \draw[xyplane=1, top color = blue!30] (0.21,1)--(1.21,1)--(1.31,1.15)--(0.31,1.15) --cycle;
            \draw[xyplane=1, top color = green!30] (0.1,1.16)--(0.25,1.16)--(0.80,1.8)--(0.65,1.8) --cycle;
            \draw[xyplane=1] node[black] at (0.17,-0.2){\(\boldsymbol{\mathrm{u_2}}\)};
            \draw[xyplane=1] node[black] at (1.3,1.34){\(\boldsymbol{\mathrm{v_2}}\)};
            \draw[xyplane=1] node[black] at (0.8,2){\(\boldsymbol{\mathrm{w_2}}\)};
        \end{scope}
        
        \begin{scope}[shift={(6.3 cm,0 cm)}]
            \draw[xyplane=2] node[black] at (0.5,1.5){\Large\(+ \dots +\)};
        \end{scope}
         \begin{scope}[shift={(7.2 cm,0 cm)}]
            \draw[xyplane=1] (0,0)--(0.15,0)--(0.15,1)--(0,1) --cycle;
            \draw[xyplane=1, top color = blue!30] (0.21,1)--(1.21,1)--(1.31,1.15)--(0.31,1.15) --cycle;
            \draw[xyplane=1, top color = green!30] (0.1,1.16)--(0.25,1.16)--(0.80,1.8)--(0.65,1.8) --cycle;
            \draw[xyplane=1] node[black] at (0.17,-0.2){\(\boldsymbol{\mathrm{u_R}}\)};
            \draw[xyplane=1] node[black] at (1.3,1.34){\(\boldsymbol{\mathrm{v_R}}\)};
            \draw[xyplane=1] node[black] at (0.8,2){\(\boldsymbol{\mathrm{w_R}}\)};
        \end{scope}
    \end{tikzpicture}
\caption{CP Factorization.}
\label{fig_cp}
\end{figure}
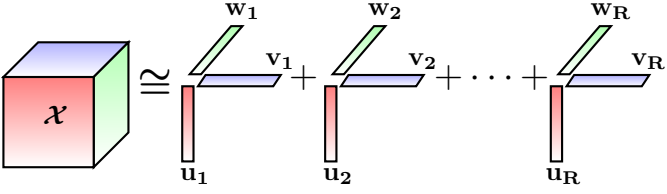

As shown in  Fig.\ref{fig_cp}, it decomposes the tensor  \(\boldsymbol{\mathcal{X}}\) into three factor matrices \(\boldsymbol{\mathrm{U}},\boldsymbol{\mathrm{V}}\) and \(\boldsymbol{\mathrm{W}}\) with \(R\) hidden features.
It is solved as the following minimization problem:
\begin{equation}
\label{eq_6}
    \min_{\boldsymbol{\mathrm{U}},\boldsymbol{\mathrm{V}}, \boldsymbol{\mathrm{W}}}f(\boldsymbol{\mathrm{U}},\boldsymbol{\mathrm{V}}, \boldsymbol{\mathrm{W}}) =  \norm{\boldsymbol{\mathcal{X}} - \llbracket\boldsymbol{\mathrm{U}}, \boldsymbol{\mathrm{V}}, \boldsymbol{\mathrm{W}} \rrbracket }^2.
    %(\boldsymbol{\mathrm{U}},\boldsymbol{\mathrm{V}}, \boldsymbol{\mathrm{W}})
\end{equation}

%\begin{figure}[!t]
%\centering
%  \begin{tikzpicture}
%  \tikzset{xzplane/.style={canvas is xz plane at y=#1, thick,shading=axis,shading angle=90,top color = blue!30,draw=black}};
%  \tikzset{yzplane/.style={canvas is yz plane at x=#1,thick,shading=axis,shading angle=90,top color = green!30,draw=black}};
%  \tikzset{xyplane/.style={canvas is xy plane at z=#1,thick,shading=axis,shading angle=90,top color = red!50,draw=black}};
%  \tikzset{xyplane2/.style={canvas is xy plane at z=#1,thick,shading=axis,shading angle=90,top color = orange!30,draw=black}};
%
%
%  \draw[xyplane2=2] (0,0)--(2,0)--(2,2)--(0,2) --cycle;
%  \draw[xyplane=2] node[black] at (1,1){\Large\(\boldsymbol{\mathrm{Y}}\)};
%
%
%  \begin{scope}[shift={(2 cm,0 cm)}]
%  
%
%  \draw[xzplane=2] (0,0)--(2,0)--(2,2)--(0,2) --cycle;
%  \draw[yzplane=2] (0,0)--(2,0)--(2,2)--(0,2) --cycle;
%  \draw[xyplane=2] (0,0)--(2,0)--(2,2)--(0,2) --cycle;
%  \draw[xyplane=2] node[black] at (1,1){\Large\(\boldsymbol{\mathcal{X}}\)};
%  \draw[dashed,color = red] (-0.78,0) --  (-0.78,-2) node[anchor=west]{$users$};
%  \draw[dashed,color = red] (-0.78,1.25) -- (3,1.25) node[anchor=west]{$items$};
%  \draw[dashed,color = red] (-0.78,1.25) -- (1.2,3.2) node[anchor=west]{$tags$};
%  \draw[dashed,color = red] (-0.78,1.25) -- (-4,1.25) node[anchor=north]{$users$};
%  \end{scope}
%  \end{tikzpicture}
%\caption{Third order tensor and matrix coupled in mode 1.}
%\label{fig2}
%\end{figure}

\textbf{Coupled Matrix Tensor Factorization:} The objective function of CMTF optimization with a third order tensor \(\boldsymbol{\mathcal{X}} \in \R^{(J\times K\times L)} \)  and matrix \(\boldsymbol{\mathrm{Y}} \in \R^{(J \times M)} \) coupled (or shared) in the first mode can be formulated \cite{acar2011all} as:
\begin{multline}
\label{eq_7}
\min_{\boldsymbol{\mathrm{U}}^{(1)},\boldsymbol{\mathrm{V}}, \boldsymbol{\mathrm{W}}, \boldsymbol{\mathrm{U}}^{(2)}}f(\boldsymbol{\mathrm{U}}^{(1)},\boldsymbol{\mathrm{V}}, \boldsymbol{\mathrm{W}}, \boldsymbol{\mathrm{U}}^{(2)}) = 
\norm{\boldsymbol{\mathcal{X}} - \llbracket\boldsymbol{\mathrm{U}}^{(1)}, \boldsymbol{\mathrm{V}}, \boldsymbol{\mathrm{W}} \rrbracket }^2 \\
+ \left\lVert\boldsymbol{\mathrm{Y}} - \boldsymbol{\mathrm{U}}^{(1)} \boldsymbol{\mathrm{U}}^{(2)}\right\rVert^2
   % (\boldsymbol{\mathrm{U}}^{(1)},\boldsymbol{\mathrm{V}},\boldsymbol{\mathrm{W}},\boldsymbol{\mathrm{U}}^{(2)})
\end{multline}
where \(\boldsymbol{\mathcal{X}}\) is factorized as a three dimensional CP model~\cite{carroll1970,kolda2009tensor} and \(\boldsymbol{\mathrm{Y}}\) is factorized as a NMF model~\cite{lee2001algorithms}. The factor matrix \(\boldsymbol{\mathrm{U}}^{(1)}  \in \R^{(J \times R)}\) is shared between \(\boldsymbol{\mathcal{X}}\) and \(\boldsymbol{\mathrm{Y}}\), whereas factor matrices \(\boldsymbol{\mathrm{V}} \in \R^{(K \times R)}\), \(\boldsymbol{\mathrm{W}} \in \R^{(L \times R)}\)  and \(\boldsymbol{\mathrm{U}}^{(2)} \in \R^{(M \times R)}\)  are unshared.

%==============
\section{The Proposed Cut-off Coordinate Descent Algorithm for N-CMTF: Cut-CD}
\begin{figure}[!t]
\centering
\includegraphics [width=2.6in, height = 3in]{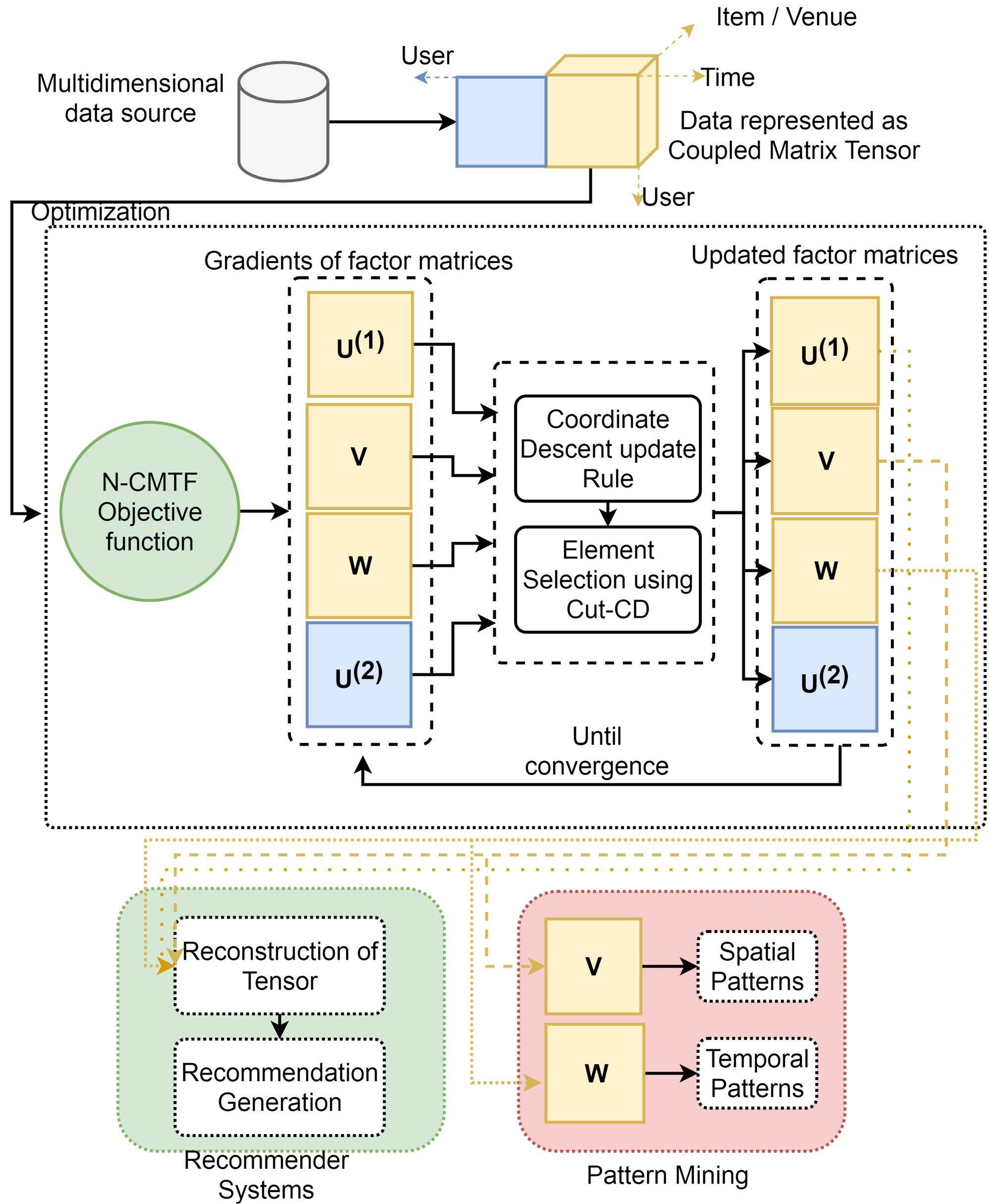}
\caption{Architecture of the overall process of Cut-CD.}
\label{arch}
\end{figure}

The overall process of Cut-CD is presented in Fig.~\ref{arch}. %N-CMTF is an optimization problem solved using a factorization algorithm. The objective is to identify the factor matrices representing each mode of the input tensor. These factor matrices when multiplied as per \eqref{eq_5} will yield the approximated tensor that should be equivalent to the input tensor. 

%-------------
\subsection{Cut-CD Objective Function: N-CMTF}
The goal of Cut-CD is to identify the factor matrices representing each mode of the input tensor and the auxiliary matrix that can approximate the input tensor and the auxiliary matrix together. The objective of the optimization problem is to minimize the Euclidean distance between the input tensor and the approximated tensor while also minimizing the Euclidean distance between the auxiliary matrix and the approximated matrix using the derived factor matrices. The objective function of CMTF as formulated in \eqref{eq_7} can be represented as the sum of two objective functions representing the tensor and matrix as:
\begin{equation}
\label{eq_8}
    %f = f_1(\boldsymbol{\mathrm{U}}^{(1)},\boldsymbol{\mathrm{V}},\boldsymbol{\mathrm{W}}) + f_2(\boldsymbol{\mathrm{U}}^{(1)}, \boldsymbol{\mathrm{U}}^{(2)})
f = f_1 + f_2,
\end{equation}
where $f_1$ is solved as,
\begin{equation}
\min_{\boldsymbol{\mathrm{U}}^{(1)},\boldsymbol{\mathrm{V}}, \boldsymbol{\mathrm{W}} \geq 0}f_1(\boldsymbol{\mathrm{U}}^{(1)},\boldsymbol{\mathrm{V}}, \boldsymbol{\mathrm{W}}) = 
\norm{\boldsymbol{\mathcal{X}} - \llbracket\boldsymbol{\mathrm{U}}^{(1)}, \boldsymbol{\mathrm{V}}, \boldsymbol{\mathrm{W}} \rrbracket }^2
\end{equation}
and $f_2$ is solved as,
\begin{equation}
\min_{\boldsymbol{\mathrm{U}}^{(1)},\boldsymbol{\mathrm{U}}^{(2)} \geq 0} f_2(\boldsymbol{\mathrm{U}}^{(1)},\boldsymbol{\mathrm{U}}^{(2)}) = \left\lVert\boldsymbol{\mathrm{Y}} - \boldsymbol{\mathrm{U}}^{(1)} \boldsymbol{\mathrm{U}}^{(2)}\right\rVert^2
\end{equation}

%-----------
\subsection{Gradient Calculation}
As the CMTF optimization is a non-convex optimization problem, the values of all other factor matrices need to be fixed to update one factor matrix at a time~\cite{acar2011all}. To update the factor matrix using the CD method~\cite{cicho2009fhals,wright2015coordinate}, we need to calculate the gradients of the factor matrices. 

We first explain the process of learning the factor matrix \(\boldsymbol{\mathrm{U}}^{(1)}\) which is a shared factor matrix during the factorization process. The gradients \(\boldsymbol{\mathrm{G}}\) for the objective function $f$ with respect to \(\boldsymbol{\mathrm{U}}^{(1)}\) is solved using partial derivatives as:
\begin{equation}
\label{eq_9}
 \boldsymbol{\mathrm{G}} = \frac{\partial f}{\partial \boldsymbol{\mathrm{U}}^{(1)}} = \frac{\partial f_1}{\partial \boldsymbol{\mathrm{U}}^{(1)}} + \frac{\partial f_2}{\partial \boldsymbol{\mathrm{U}}^{(1)}}  
\end{equation}

\begin{equation}
    \label{eq_10}
    \frac{\partial f_1}{\partial \boldsymbol{\mathrm{U}}^{(1)}} = 
    -\boldsymbol{\mathrm{X_1}}(\boldsymbol{\mathrm{W}} \odot \boldsymbol{\mathrm{V}})
    + \boldsymbol{\mathrm{U}}^{(1)}(\boldsymbol{\mathrm{V}}^T \boldsymbol{\mathrm{V}} \ast \boldsymbol{\mathrm{W}}^T \boldsymbol{\mathrm{W}})
\end{equation}
\begin{equation}
    \label{eq_11}
    \frac{\partial f_2}{\partial \boldsymbol{\mathrm{U}}^{(1)}} = 
     \boldsymbol{\mathrm{U}^{(1)}} \boldsymbol{\mathrm{U}}^{(2)^T} \boldsymbol{\mathrm{U}}^{(2)} 
     - \boldsymbol{\mathrm{Y}}\boldsymbol{\mathrm{U}}^{(2)}
\end{equation}
where \(\boldsymbol{\mathrm{X_1}}\) is the mode-1 matricization of the tensor. \eqref{eq_9} can be rewritten using \eqref{eq_10} and \eqref{eq_11} as follows:
\begin{multline}
\label{eq_12}
    \boldsymbol{\mathrm{G}} = 
       -\boldsymbol{\mathrm{X_1}}(\boldsymbol{\mathrm{W}} \odot \boldsymbol{\mathrm{V}})
    + \boldsymbol{\mathrm{U}}^{(1)}(\boldsymbol{\mathrm{V}}^T \boldsymbol{\mathrm{V}} \ast \boldsymbol{\mathrm{W}}^T \boldsymbol{\mathrm{W}}) \\ 
    + \boldsymbol{\mathrm{U}^{(1)}} \boldsymbol{\mathrm{U}}^{(2)^T} \boldsymbol{\mathrm{U}}^{(2)} 
    -\boldsymbol{\mathrm{Y}}\boldsymbol{\mathrm{U}}^{(2)}.
\end{multline}

For simplicity, each element of the gradient $\boldsymbol{\mathrm{G}}$ is represented as:
\begin{multline}
\label{eq_13}
g_{jr} = 
       - (\boldsymbol{\mathrm{X_1}}(\boldsymbol{\mathrm{W}} \odot \boldsymbol{\mathrm{V}}))_{jr}
    + (\boldsymbol{\mathrm{U}}^{(1)}(\boldsymbol{\mathrm{V}}^T \boldsymbol{\mathrm{V}} \ast \boldsymbol{\mathrm{W}}^T \boldsymbol{\mathrm{W}}))_{jr} \\ 
    + (\boldsymbol{\mathrm{U}^{(1)}} \boldsymbol{\mathrm{U}}^{(2)^T} \boldsymbol{\mathrm{U}}^{(2)})_{jr} 
    - (\boldsymbol{\mathrm{Y}}\boldsymbol{\mathrm{U}}^{(2)})_{jr}
\end{multline}
where \(g_{jr}\) is the gradient of \( (j,r)^{th} \) element of \( \boldsymbol{\mathrm{U}}^{(1)} \).

The second-order partial derivative $\boldsymbol{\mathrm{H}}$ of function $f$ with respect to $\boldsymbol{\mathrm{U}}^{(1)}$ is defined as,
\begin{equation}
\label{eq_15}
\boldsymbol{\mathrm{H}} = \frac{\partial^2 f}{\partial \boldsymbol{\mathrm{U}}^{(1)}} = (\boldsymbol{\mathrm{V}}^T\boldsymbol{\mathrm{V}} \ast \boldsymbol{\mathrm{W}}^T\boldsymbol{\mathrm{W}}) + \boldsymbol{\mathrm{U}}^{(2)^T}\boldsymbol{\mathrm{U}}^{(2)}.
\end{equation}

For simplicity, each element of the second-order partial derivative $\boldsymbol{\mathrm{H}}$ for each element can be represented as:
\begin{equation}
\label{eq_16}
h_{rr} = (\boldsymbol{\mathrm{V}}^T\boldsymbol{\mathrm{V}} \ast \boldsymbol{\mathrm{W}}^T\boldsymbol{\mathrm{W}})_{rr} + (\boldsymbol{\mathrm{U}}^{(2)^T}\boldsymbol{\mathrm{U}}^{(2)})_{rr}.
\end{equation}
%-----------
\subsection{Coordinate Descent based Update Rule}
The CD update rule can be given as:
\begin{equation}
\label{eq_14}
    \Hat{\boldsymbol{\mathrm{U}}}^{(1)} = \max\left(0,\left(\boldsymbol{\mathrm{U}}^{(1)}-\frac{\boldsymbol{\mathrm{G}}}{\boldsymbol{\mathrm{H}}}\right)\right) - \boldsymbol{\mathrm{U}}^{(1)}. 
\end{equation}
%where \( \boldsymbol{\mathrm{H}} \) is the Hadamard product calculated as the second-order partial derivative of the objective function:

The single element update rule of CD is given as \cite{cicho2009fhals}: 
\begin{equation}
    \label{eq_17}
    \Hat{u}^{(1)}_{jr} = \max\left(0,\left(u^{(1)}_{jr} - \frac{g_{jr}}{h_{rr}}\right)\right) - u^{(1)}_{jr}
\end{equation}
where \( \Hat{u}^{(1)}_{jr} \) indicates the computed element, and \( u^{(1)}_{jr} \) indicates the \( (j,r)^{th} \) element of the factor matrix \( \boldsymbol{\mathrm{U}}^{(1)} \), 
\begin{equation}
\label{eq_18}
    u^{(1)}_{jr} \xleftarrow{} u^{(1)}_{jr} + \Hat{u}^{(1)}_{jr}.
\end{equation}

%--------------
\subsection{Nonnegative Constraint}
We impose the nonnegative constraint to the element as:
\begin{equation}
 u^{(1)}_{jr}  =
  \begin{cases}
    u^{(1)}_{jr}      & \quad \text{for } u^{(1)}_{jr} > 0\\
    0  & \quad \text{for } u^{(1)}_{jr} \leq 0.
  \end{cases}
\end{equation}

%------------
\subsection{Cut-off Coordinate Descent (Cut-CD) – Column-wise Element Selection}
The calculations of \(\boldsymbol{\mathrm{X_1}} (\boldsymbol{\mathrm{W}} \odot \boldsymbol{\mathrm{V}}) \),  \(\boldsymbol{\mathrm{V}}^T\boldsymbol{\mathrm{V}} \ast \boldsymbol{\mathrm{W}}^T\boldsymbol{\mathrm{W}}\) and \(\boldsymbol{\mathrm{U}^{(1)}} \boldsymbol{\mathrm{U}}^{(2)^T} \boldsymbol{\mathrm{U}}^{(2)}\) for every element are expensive. The computational complexity of a factorization algorithm can be reduced if these values do not need to be calculated for every element and they are only calculated for selected elements. We now present how the elements can be selected and updated using the proposed Cut-CD algorithm.

We conjecture that the N-CMTF algorithm will converge faster by updating the important elements repeatedly, instead of updating all elements sequentially. Algorithm 1  details the process. We propose to calculate the importance of an element during factorization using the gradient principle \cite{gcd2011} as,
\begin{equation}
\label{eq_20}
    e_{jr} = -(u^{(1)}_{jr} \ast g_{jr}) - 0.5(h_{rr}\ast u^{(1)}_{jr} \ast u^{(1)}_{jr})
\end{equation}
where \(e_{jr}\) is the \((j,r)^{th}\) element importance which is the difference in objective function. The higher the value, the higher the importance.

We propose to calculate the column-wise element importance as,
\begin{equation}
\label{eq_21}
    \boldsymbol{\mathrm{e_{*r}}} = -(\boldsymbol{\mathrm{u^{(1)}_{*r}}} \ast \boldsymbol{\mathrm{g_{*r}}}) - 0.5(\boldsymbol{\mathrm{h_{*r}}}\ast \boldsymbol{\mathrm{u^{(1)}_{*r}}} \ast \boldsymbol{\mathrm{u^{(1)}_{*r}}})
\end{equation}
where \(\boldsymbol{\mathrm{e_{*r}}}\) is the vector of element importance for all the elements in \(r^{th}\) column. 

When an element \(u^{(1)}_{jr}\) is updated using a traditional element selection-based algorithm such as GCD \cite{gcd2011}, the gradient \(\boldsymbol{\mathrm{g_{j*}}}\) and the element importance \(\boldsymbol{\mathrm{e_{j*}}}\) of the entire row are need to be calculated for each update. The calculation of gradient and element importance for each update is an expensive task and consumes additional runtime and memory. Additionally, the gradients are calculated repeatedly for all elements that are present in the same row where many of them are not required. Due to this reason, we propose to update them column-wise so that the repeated updates can be avoided, and the efficiency of the factorization can be improved.
Once we have the vector \(\boldsymbol{\mathrm{e_{*r}}}\), the element values are normalized as:
\begin{equation}
    \label{eq_22}
    \boldsymbol{\mathrm{n_{*r}}} = \frac{\boldsymbol{\mathrm{e_{*r}}}-\min(\boldsymbol{\mathrm{e_{*r}})}}{\max(\boldsymbol{\mathrm{e_{*r}}})-\min(\boldsymbol{\mathrm{e_{*r}}})}
\end{equation}
where \(\boldsymbol{\mathrm{n_{*r}}}\) indicates the normalized column \(r\) of \(\boldsymbol{\mathrm{e_{*r}}}\).

Instead of updating a single element repeatedly, a set of elements in each column based on a cut-off value can be updated only once. Most importantly, for each iteration, we propose to utilise a pre-calculated gradient that can help in identifying the importance of each element in each column. Therefore, we do not need to perform any inner gradient updates which are the bottleneck in row-wise element updates of a CD-based CMTF algorithm \cite{gcd2011}. 
\begin{figure}[htb!]
    \centering
    \subfloat[GCD]{{
     \includegraphics[width=3.7cm, height=2.5cm]{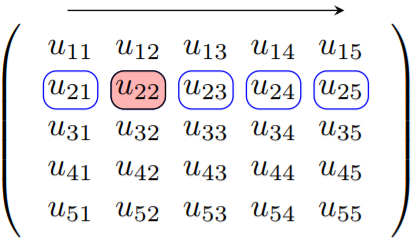} 
    }}%
    \qquad
    \subfloat[The Proposed Cut-CD]{{
     \includegraphics[width=3.7cm, height=2.5cm]{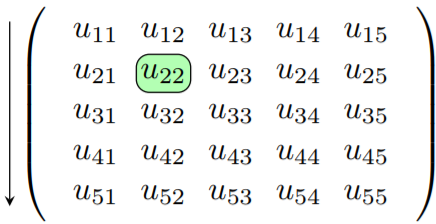} 
    }}%
    \caption{(a) Row-wise (GCD) and (b) Colum-wise (Cut-CD) element selection in a factor matrix; red color indicates the calculation for an element, element importance and gradient updates; blue color indicates the calculation for element importance and gradient updates; green color indicates the element update calculation alone.}%
    \label{fig:cutcd}%
\end{figure}

As shown in Fig.~\ref{fig:cutcd}a, if element $u_{22}$ is selected to update in GCD, it is essential to update the gradients of the entire row $\boldsymbol{\mathrm{u_{2\ast}}}$ and the element importance $\boldsymbol{\mathrm{e_{2\ast}}}$. This is required to choose the next important element in the row for an update. The same process is repeated for every row and consequently increases the time complexity.   

We propose to choose \(mean(\boldsymbol{\mathrm{n_{*r}}} )\) as the cut-off value for each column as the mean value represents the best cut-off value (as shown in the experiment section) and avoids the need of parameter tuning. As shown in Fig.~\ref{fig:cutcd}b, if element $u_{22}$ is selected to update in Cut-CD, it is enough to update the element alone and to avoid gradient updates, a distinct cut-off value can be set for each column. Since it does not need to be dependent on other columns of $\boldsymbol{\mathrm{G}}$ thereby gradient updates are minimized. Additionally, since each element is updated only once in an iteration, it is not required to calculate element importance repeatedly. This is supported by the following lemma 4.1.

\RestyleAlgo{boxruled}
\LinesNumbered
\begin{algorithm}
\SetAlgoLined
\textbf{Input:} Tensor \( \boldsymbol{\mathcal{X}} \); Matrix \( \boldsymbol{\mathrm{Y}} \); Randomly Initiated factor matrices \(\boldsymbol{\mathrm{U}}^{(1)} \in \R^{J \times R}\), \(\boldsymbol{\mathrm{V}} \in \R^{K \times R}\), \(\boldsymbol{\mathrm{W}} \in \R^{L \times R}\), \(\boldsymbol{\mathrm{U}}^{(2)} \in \R^{M \times R}\); Rank \( R \); \(\boldsymbol{\mathrm{E}} = \varnothing\);\(\boldsymbol{\mathrm{N}} = \varnothing\); maxiters. \\
%\textbf{Compute:} \( \boldsymbol{\mathrm{G}} \), \( \boldsymbol{\mathrm{H}} \);  using \eqref{eq_12} and \eqref{eq_15}. \\
\textbf{Result:} Learned Factor matrices \(\boldsymbol{\mathrm{U}}^{(1)}\), \(\boldsymbol{\mathrm{V}}\),\(\boldsymbol{\mathrm{W}}\),  \(\boldsymbol{\mathrm{U}}^{(2)}\) \\
\For{$iters=  1:maxiters$} 
{
\textbf{Compute:} \( \boldsymbol{\mathrm{G}} \), \( \boldsymbol{\mathrm{H}} \);  using \eqref{eq_12} and \eqref{eq_15}. \\
%  \For {each of the factor matrices \(\boldsymbol{\mathrm{U}}^{(1)}\), \(\boldsymbol{\mathrm{V}}\),\(\boldsymbol{\mathrm{W}}\),  \(\boldsymbol{\mathrm{U}}^{(2)}\)}
%{
	\For{\(r = 1:R\)} 
    {
    calculate \(\boldsymbol{\mathrm{e_{*r}}}\) using \eqref{eq_21} \;
    calculate \(\boldsymbol{\mathrm{n_{*r}}}\) using \eqref{eq_22} \;
    $\text{cut-off}$ = mean($\boldsymbol{\mathrm{n_{*r}}}$) \;
     \For{\(j = 1:J\)} 
     {
     \uIf{\(n_{jr} \geq \text{cut-off} \)}{
    update the $(j,r)^{th}$ element of $\boldsymbol{\mathrm{U^{(1)}}}$\;
  }
     }
    }
    Repeat analogues lines 4 to 13 with $\boldsymbol{\mathrm{G}}$ and $\boldsymbol{\mathrm{H}}$ calculated using \eqref{eq_23} and \eqref{eq_24} respectively to update elements of $\boldsymbol{\mathrm{V}}$ in line 11;\\
    Repeat analogues lines 4 to 13 with $\boldsymbol{\mathrm{G}}$ and $\boldsymbol{\mathrm{H}}$ calculated using \eqref{eq_25} and \eqref{eq_26} respectively to update elements of $\boldsymbol{\mathrm{W}}$ in line 11;\\
    Repeat analogues lines 4 to 13 with $\boldsymbol{\mathrm{G}}$ and $\boldsymbol{\mathrm{H}}$ calculated using \eqref{eq_27} and \eqref{eq_28} respectively to update elements of $\boldsymbol{\mathrm{U^{(2)}}}$ in line 11;
%}
}
\label{alg1}
\caption{Cut-off Coordinate Descent Cut-CD}
\end{algorithm}

\begin{lemma}
In the column-wise Cut-CD element selection, the update of elements and element importance in $r^{th}$ column of \(\boldsymbol{\mathrm{U}}^{(1)}\) is dependent only on the gradients of the \(r^{th}\) column.
\end{lemma} 
\begin{proof}
	Let \(u^{(1)}_{jr}\) $\in$ \(\boldsymbol{\mathrm{U}}^{(1)}\) be \((j,r)^{th} \) element of \(\boldsymbol{\mathrm{U}}^{(1)}\) and \(g_{jr}\) $\in$ \(\boldsymbol{\mathrm{G}}\) represents \( (j,r)^{th}\) gradient of \(\boldsymbol{\mathrm{U}}^{(1)}\).  $\Omega$ denotes the set of indices of observable entries of \( \boldsymbol{\mathcal{X}} \).\\
	
	$\forall$ ~~r: $r \in R$, the gradient can be calculated column-wise as,
	\begin{equation}
	\label{eq:lem1}
	\boldsymbol{\mathrm{g_{*r}}} = \boldsymbol{\mathrm{(X_1(W\odot V))_{*r}}} + \boldsymbol{\mathrm{(U}}^{(1)}\boldsymbol{\mathrm{(V}}^T\boldsymbol{\mathrm{V}} \ast \boldsymbol{\mathrm{W}}^T\boldsymbol{\mathrm{W))_{*r}}}.
	\end{equation}
	
	The MTTKRP operation ($\boldsymbol{\mathrm{X_1(W\odot V)}}$) for entire matrix is an expensive task which can be simplified by calculating it column-wise as,
	
	\begin{equation}
	\label{eq:lem2}
	 \boldsymbol{\mathrm{(X_1(W\odot V))_{*r}}} = \boldsymbol{\mathrm{m_{*r}}} = \sum_{(j,k,l) \in \Omega^{U^{(1)}_j}}(x_{jkl}*(\boldsymbol{\mathrm{w_{*r}}} \boldsymbol{\mathrm{v_{*r}}}))
	\end{equation}
	where $\Omega^{U^{(1)}_j}$ indicates the subset of $\Omega$ whose mode $\boldsymbol{\mathrm{U^{(1)}}}$'s index is $j$. $\boldsymbol{\mathrm{w_{*r}}}$ and $\boldsymbol{\mathrm{v_{*r}}}$ indicate the $r^{th}$ column of the factor matrices $\boldsymbol{\mathrm{W}}$ and $\boldsymbol{\mathrm{V}}$ respectively.
	
	Substituting \eqref{eq:lem2} in \eqref{eq:lem1}, we have,
		\begin{equation}
	\label{eq:lem3}
	\boldsymbol{\mathrm{g_{*r}}} = \boldsymbol{\mathrm{m_{*r}}} + \boldsymbol{\mathrm{(U}}^{(1)}\boldsymbol{\mathrm{(V}}^T\boldsymbol{\mathrm{V}} \ast \boldsymbol{\mathrm{W}}^T\boldsymbol{\mathrm{W))_{*r}}}.
	\end{equation}

	As the calculation of column-wise gradient is possible as per \eqref{eq:lem3}, the next ($(r+1)^{th}$) column's gradient is calculated as, 
	\begin{equation}
	\label{eq:lem4}
	\boldsymbol{\mathrm{g_{*r+1}}} = \boldsymbol{\mathrm{m_{*r+1}}} + \boldsymbol{\mathrm{(U}}^{(1)}\boldsymbol{\mathrm{(V}}^T\boldsymbol{\mathrm{V}} \ast \boldsymbol{\mathrm{W}}^T\boldsymbol{\mathrm{W))_{*r+1}}}.
	\end{equation}
	
	From \eqref{eq:lem3} and \eqref{eq:lem4}, it can be seen that calculating the gradient of one column is independent of the gradient of the other column. Moreover, the column-wise element importance as per \eqref{eq_21} is dependent only on the $r^{th}$ column of the gradient ($\boldsymbol{\mathrm{g_{*r}}}$).
	
	The set of important elements $\boldsymbol{\mathrm{s_r}} \in \boldsymbol{\mathrm{u^{(1)}_{*r}}} \leq$ \textit{cut-off} are selected. This subset $\boldsymbol{\mathrm{s_r}}$ is dependent on the element importance calculated using \eqref{eq_21} which in-turn dependent on the $r^{th}$ column of $\boldsymbol{\mathrm{G}}$ as per \eqref{eq:lem1}. 
	Hence, the elements and element importance of $r^{th}$ column of $\boldsymbol{\mathrm{U}^{(1)}}$ can be calculated and updated column-wise without depending on the gradients of the other columns. 
	
	%The number of gradients calculated for each column is the total number of rows. The element selection using Cut-CD selects only a subset of it to update in each column.
	
	If $J$ is the total number of rows, Cut-CD selects $S$ (where $S< J$) elements based on the distinct cut-off value, to update in each column. This considerably minimizes the total number of element updates.
	
	Therefore, for each update of \(u^{(1)}_{jr}\), it is enough to update \(g_{jr}\) alone without updating \(\boldsymbol{\mathrm{g_{j*}}}\) of all the columns of \(j^{th}\) row and element importance \(e_{jr}\).
	\end{proof}
%\begin{IEEEproof}
\begin{comment}
\begin{proof}  Let \(u^{(1)}_{jr}\) $\in$ \(\boldsymbol{\mathrm{U}}^{(1)}\) be \((j,r)^{th} \) element of \(\boldsymbol{\mathrm{U}}^{(1)}\) and \(g_{jr}\) $\in$ \(\boldsymbol{\mathrm{G}}\) represents \( (j,r)^{th}\) gradient of \(\boldsymbol{\mathrm{U}}^{(1)}\).\\
For every \(r:R\), the set of important elements $\boldsymbol{\mathrm{s_r}} \in \boldsymbol{\mathrm{u_{*r}}}^{(1)} \leq$ \textit{cut-off} are selected. \\  
The subset \(\boldsymbol{\mathrm{s_r}}\) is dependent on \(r^{th}\) column of \(\boldsymbol{\mathrm{G}}\).\\
For each update of \(u^{(1)}_{jr}\) $\in$ \(\boldsymbol{\mathrm{s_r}}\), \(g_{jr}\)  alone updated as \(\boldsymbol{\mathrm{s_r}}\) $\independent{}$ \(\boldsymbol{\mathrm{g_{*p}}}\) where \(p \neq r\) indicates all other columns except \(r\) and $\independent{}$ indicates that \(\boldsymbol{\mathrm{s_r}}\) is independent of \(\boldsymbol{\mathrm{g_{*p}}}\).\\
Therefore, for each update of \(u^{(1)}_{jr}\), it is enough to update \(g_{jr}\) alone and it is not necessary to update \(\boldsymbol{\mathrm{g_{j*}}}\) of all the columns of \(j^{th}\) row and element importance \(e_{jr}\).
%\end{IEEEproof}
\end{proof}
\end{comment}

As the matrix component \( \boldsymbol{\mathrm{Y}} \) is shared only with mode-1 (\(\boldsymbol{\mathrm{U}}^{(1)}\)), the partial derivatives of \(f_2\) in \eqref{eq_9} can be set to zero for updating the tensor factor matrices \(\boldsymbol{\mathrm{V}}\),\(\boldsymbol{\mathrm{W}}\). The update rule for \(\boldsymbol{\mathrm{V}}\) and \(\boldsymbol{\mathrm{W}}\) can be derived by the following gradient \(\boldsymbol{\mathrm{G}}\) and second-order partial derivative \(\boldsymbol{\mathrm{H}}\) calculations.
\begin{multline}
\label{eq_23}
    \frac{\partial f}{\partial \boldsymbol{\mathrm{V}}} = \boldsymbol{\mathrm{G}} = 
       -\boldsymbol{\mathrm{X_2}}(\boldsymbol{\mathrm{W}} \odot \boldsymbol{\mathrm{U}}^{(1)})
    + \boldsymbol{\mathrm{V}}(\boldsymbol{\mathrm{U}}^{(1)^T} \boldsymbol{\mathrm{U}}^{(1)} \ast \boldsymbol{\mathrm{W}}^T \boldsymbol{\mathrm{W}}).
\end{multline}
\begin{equation}
   \label{eq_24}
    \frac{\partial^2 f}{\partial \boldsymbol{\mathrm{V}}} = \boldsymbol{\mathrm{H}} = 
       \boldsymbol{\mathrm{U}}^{(1)^T} \boldsymbol{\mathrm{U}}^{(1)} \ast \boldsymbol{\mathrm{W}}^T \boldsymbol{\mathrm{W}}.
 \end{equation}
\begin{multline}
\label{eq_25}
    \frac{\partial f}{\partial \boldsymbol{\mathrm{W}}} = \boldsymbol{\mathrm{G}} = 
       -\boldsymbol{\mathrm{X_3}}(\boldsymbol{\mathrm{V}} \odot \boldsymbol{\mathrm{U}}^{(1)})
    + \boldsymbol{\mathrm{W}}(\boldsymbol{\mathrm{U}}^{(1)^T} \boldsymbol{\mathrm{U}}^{(1)} \ast \boldsymbol{\mathrm{V}}^T \boldsymbol{\mathrm{V}}).
\end{multline}
\begin{equation}
   \label{eq_26}
    \frac{\partial^2 f}{\partial \boldsymbol{\mathrm{W}}} = \boldsymbol{\mathrm{H}} = 
       \boldsymbol{\mathrm{U}}^{(1)^T} \boldsymbol{\mathrm{U}}^{(1)} \ast \boldsymbol{\mathrm{V}}^T \boldsymbol{\mathrm{V}}.
 \end{equation}
 
When updating \(\boldsymbol{\mathrm{U}}^{(2)}\) in \eqref{eq_8}, the partial derivative of \(f_1\) with respect to tensor in \eqref{eq_9} is set to zero, the gradient \(\boldsymbol{\mathrm{G}}\) and second-order partial derivative \(\boldsymbol{\mathrm{H}}\) for updating \(\boldsymbol{\mathrm{U}}^{(2)}\)  becomes,
\begin{multline}
\label{eq_27}
    \frac{\partial f}{\partial \boldsymbol{\mathrm{U}}^{(2)}} = \boldsymbol{\mathrm{G}} = 
       \boldsymbol{\mathrm{U}}^{(2)}\boldsymbol{\mathrm{U}}^{(1)^T} \boldsymbol{\mathrm{U}}^{(1)} - \boldsymbol{\mathrm{Y}}^T\boldsymbol{\mathrm{U}}^{(2)}.
\end{multline}
\begin{equation}
   \label{eq_28}
    \frac{\partial^2 f}{\partial \boldsymbol{\mathrm{U}}^{(2)}} = \boldsymbol{\mathrm{H}} = 
       \boldsymbol{\mathrm{U}}^{(1)^T} \boldsymbol{\mathrm{U}}^{(1)}.
 \end{equation}
 
The Cut-CD column-wise element selection is applied to select and update the elements of factor matrices \(\boldsymbol{\mathrm{V}}\),\(\boldsymbol{\mathrm{W}}\),\(\boldsymbol{\mathrm{U}}^{(2)}\) using \eqref{eq_18} in the same way as explained for \(\boldsymbol{\mathrm{U}}^{(1)}\).
 
Algorithm 1  details the process. Cut-CD is generic and can be easily derived for NMF or NTF problems by substituting zero for \(f_1\) and \(f_2\) respectively in \eqref{eq_8} or adding another shared matrix in other tensor dimensions.

The element selection method of GCD needs to know the values of gradients of the whole matrix. This is not convenient when the algorithm is applied to a distributed or parallel environment. Whereas, the column-wise element selection as in Cut-CD allows the algorithm to efficiently adapt to a distributed or parallel environment. This is due to the fact that the element selection is independent of other columns of a factor matrix as proven in Lemma 4.1. The column-wise update sequence as proposed in PCMF and CDTF has huge advantages in a distributed and parallel environment over Cut-CD. However, this column-wise update sequence with distributed and parallelization techniques of PCMF and CDTF can be adapted to Cut-CD to make it more efficient in a distributed and parallel environment. This is beyond the scope of this paper.  
\section{Theoretical Analysis}
In this section, we will analyse Cut-CD in terms of convergence, time complexity, memory requirement and element/gradient update. As GCD is the state-of-the-art element selection-based method, most of our analytical comparison is with GCD. We use the following symbols in the analysis, \(R\): rank; \(Z\): number of element updates; and \(J,K,L\) and \(M\) indices for factor matrices \(\boldsymbol{\mathrm{U}}^{(1)}\),\(\boldsymbol{\mathrm{V}}\),\(\boldsymbol{\mathrm{W}}\) and \(\boldsymbol{\mathrm{U}}^{(2)}\) respectively.

%------------
\subsection{Convergence Analysis}
The class of optimization methods solving non-convex optimization by alternating between multiple sets of variables is known as Alternating Direction Method of Multipliers (ADMM)~\cite{boyd2011distributed}. All CD methods including Cut-CD employ ADMM in solving the optimization problem. 

\begin{theorem}
	The ADMM method with an update sequence using element section based Cut-CD converges to a stationary point for a cut-off value imposed. %different local minima depending on the cut-off value imposed.
\end{theorem}
\begin{proof}
	The optimization function $f$ reaches a stationary point (i.e. local, global, or saddle) only when the following inequality satisfies.
	\begin{equation}
	f(u^{i+1}) \leq f(u^i)
	\end{equation}
	where \(i\) indicates the \(i^{th}\) iteration. This states that the function decreases monotonically and a stationary point is reached.
	
	ADMM based update uses the update sequence,
	\begin{equation}
	\label{eq_31}
	f(u^{i+1}) = \max(0, f(u^i) - \frac{f'(u^{i+1})}{f''(u^{i+1})}), i=0,1, \dots
	\end{equation}
	
	The gradients for the sequence updating all the elements can be defined as:
	\begin{equation}
	f'(u^i) = - \sum_{j=1}^J\sum_{r=1}^R g^i_{jr}.
	\end{equation}
	
	While $\boldsymbol{\mathrm{S}}$ is a set of elements selected based on a cut-off value ($mean(\boldsymbol{\mathrm{n_{*r}}})$) using Cut-CD to be updated in \(\boldsymbol{\mathrm{U}}^{(1)}\), the gradients for the selected elements can be defined as,
	\begin{equation}
	f'_s(u^i) = - \sum_{j,r \in \boldsymbol{\mathrm{S}}} g^i_{jr}.
	\end{equation}
	
	Since the elements are selected such that the objective function will be minimized, the following inequality exists,
	
	\begin{equation}
	\label{eq_comp}
	f'_s(u^i) \leq f'(u^i).
	\end{equation}
	
	second-order partial derivative of $f$ can be defined as,
	
	\begin{equation}
	\label{eq_33}
	f_s''(u^i) = \sum_{r = 1}^R h^i_{rr} >  \sum_{r \in \boldsymbol{\mathrm{S}}} h^i_{rr} > 0.
	\end{equation}
	
	With the calculated gradients and second-order partial derivatives, as per~\eqref{eq_31}, we have
	
	\begin{equation}
	f(u^{i+1}) \leq f(u^i) - \frac{f_s'(u^{i+1})}{f_s''(u^{i+1})}.
	\end{equation} 
	
	Iterating the inequality above gives,
	
	\begin{equation}
	\sum_{i=0}^{\infty}\frac{f_s'(u^{i+1})}{f_s''(u^{i+1})} \leq f_s(u^i) \leq f(u^i)
	\end{equation}
	which proves that for the given cut-off value ($mean(\boldsymbol{\mathrm{n_{*r}}})$), $f(u^{i+1}) \leq f(u^i)$, hence Cut-CD will converge to a stationary point. 
			\end{proof}
	If Cut-CD is stuck in a saddle point, a second-order derivative test can be used to identify it and an advanced gradient descent techniques like perturbed gradient descent~\cite{jin2017escape} can be used to escape. 
\subsection{Time Complexity Analysis}
The time complexity of Cut-CD is \(O(JKLR+2JR^2+JMR+Z)\) in comparison to the time complexity of GCD \(O(JKLR+3JR^2+JMR+3ZR)\) that is $O(JR^2 + Z(1-3R))$ less than GCD.

For the first iteration, both Cut-CD and GCD need  gradients to be initialized \(\boldsymbol{\mathrm{G}}\), as shown in \eqref{eq_12}. This step includes calculation of four terms \(\boldsymbol{\mathrm{X_1}}(\boldsymbol{\mathrm{W}} \odot \boldsymbol{\mathrm{V}}),\boldsymbol{\mathrm{U}}^{(1)}(\boldsymbol{\mathrm{V}}^T \boldsymbol{\mathrm{V}} \ast \boldsymbol{\mathrm{W}}^T \boldsymbol{\mathrm{W}}), \boldsymbol{\mathrm{U}^{(1)}} \boldsymbol{\mathrm{U}}^{(2)^T} \boldsymbol{\mathrm{U}}^{(2)}\) and \(\boldsymbol{\mathrm{Y}}\boldsymbol{\mathrm{U}}^{(2)}\) which incur the complexity of \(O(JKLR),  O(JR^2), O(JR^2)\) and \(O(JMR)\) respectively. Additionally, GCD needs initialization of the element importance matrix \(\boldsymbol{\mathrm{E}}\) that will exhibit \(O(JR^2)\). If \(Z\) number of elements are updated for every inner loop, the time computation of element update costs \(O(ZR)\). Similarly, for each gradient and element importance matrix \(\boldsymbol{\mathrm{E}}\) update, it requires \(O(ZR)\). On the other hand, as shown in Algorithm 1, Cut-CD calculates the element importance column-wise to avoid initialization. Also as per lemma 4.1, Cut-CD doesn’t need to update the gradient and importance matrix \(\boldsymbol{\mathrm{E}}\) of entire row and hence the time complexity becomes \(O(Z)\). For each factor matrix update, GCD involves the $R$ inner loops, hence the time complexity of GCD and Cut-CD is \(O(JR^2)\) and \(O(JR)\) respectively.  In Table~\ref{table_complexity}, we fix the value for J as 1000 and substitute different values of R for the inner loop’s time complexity of GCD and Cut-CD. The results shows that the computational complexity of GCD grows exponentially; however, Cut-CD is efficient and grows only linearly with increase in the tensor rank.%Table~\ref{table_complexity} shows that the computational complexity of GCD grows exponentially; however,  Cut-CD is efficient and grows only linearly with increase in the tensor rank.

\begin{table}[tb!]
\caption{Time Complexity Analysis.}
\label{table_complexity}
\centering
\begin{tabular}{llllc}
\hline
Mode length (J) & Rank (R) & O(GCD) & O(Cut-CD) & Times Faster\\
\hline
1000   & 10 &  100,000  &  10,000 & 10\\
1000   & 20 &  400,000  &  20,000 & 20\\
1000   & 30 &  900,000  &  30,000 & 30\\
1000   & 50 &  2,500,000 &  50,000 & 50\\
1000   & 100 & 10,000,000 &  100,000 & 100\\
\hline
\end{tabular}
\end{table}
%------------
\subsection{Memory Requirement Analysis}
The memory requirement of Cut-CD for a single element update is \(O(1)\) in comparison to GCD that is \(O(R)\). 

For each single element update in a factor matrix, Cut-CD requires the following three types of data in the memory: (1) the element, (2) its gradient and (3) the second-order derivative value of the element. According to lemma 4.1, each element update in Cut-CD requires only one gradient update and hence the memory requirement is \(O(1)\). The independent column-wise update of Cut-CD makes it easier to perform efficiently if the process of an update can be distributed column-wise across multiple machines. On the other hand, GCD needs the gradient of the entire row to be updated for each single element update. If \(R\) is the number of columns in a factor matrix, the memory requirement is \(O(1) + O(R)\).  The memory requirement of GCD grows linearly when the rank is increased, in comparison to Cut-CD that has the same memory requirement~\(O(1)\) irrespective of the increase in rank. Importantly, it is to be noted that the selection of elements depends on the entire factor matrix, consequently, it becomes hard to process GCD in a distributed environment. 

%-----------
\subsection{Element and Gradient Update Analysis}
In CD-based factorization, the factor matrices are updated element-wise \cite{cichocki2009nonnegative}. Using a synthetic dataset, we show the total number of elements and gradients updated in each iteration by Cut-CD and GCD in Fig. 4. As proved in lemma 4.1, each element update in Cut-CD requires only one gradient update and the number of gradient updates is the same as the number of element updates. Whereas in GCD, the number of gradient updates is more than the number of element updates. Fig. 4 shows that Cut-CD does not need frequent gradient updates, as in GCD, when updating the elements in a factor matrix. It is enough to update all the gradients of a factor matrix once for each iteration in Cut-CD. It also shows that the number of element updates is higher in early iterations in GCD, whereas, the element updates count in Cut-CD remains constant. The greedy search strategy of GCD can select the same element to be updated multiple times within a single iteration while the threshold based search strategy of Cut-CD selects and updates a set of elements only once. 
\begin{figure}[ht!]
	\centering
	\subfloat{{\includegraphics[width=5.3cm,height=3.8cm]{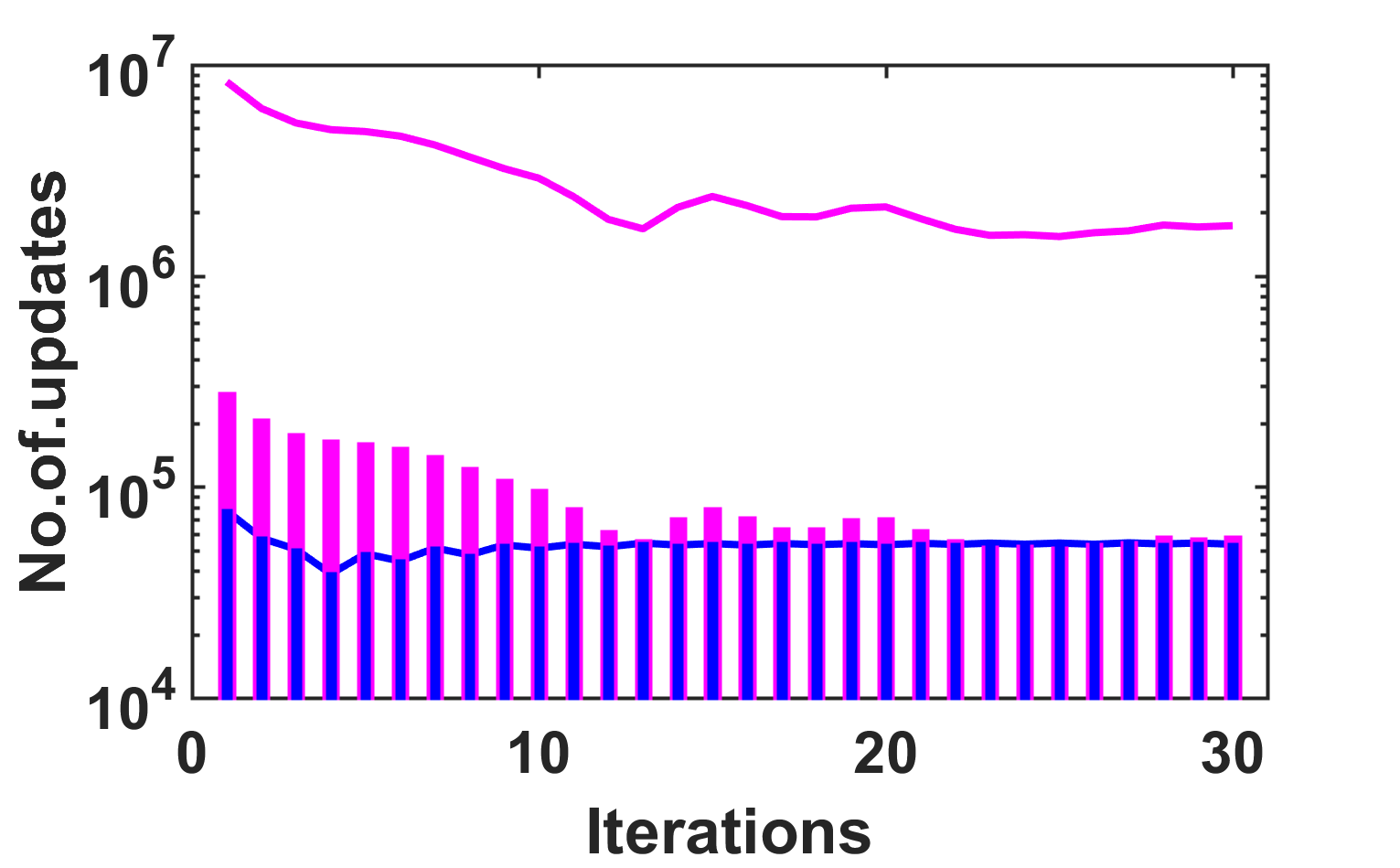}}}%
	%\qquad	
	\subfloat{\raisebox{11ex}{\includegraphics[width=0.9in, height=0.4in]{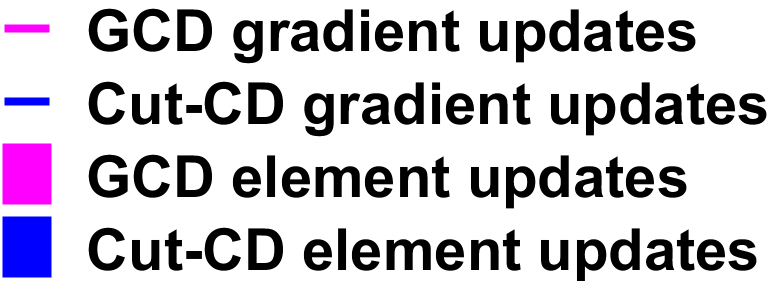}}}%
	\caption{A factor matrix update analysis on the Syn1 dataset (as shown in Table~\ref{table_dataset}).}%
	\label{fig:updateanalysis}
\end{figure}
\section{Empirical Analysis}
We conducted experiments with Cut-CD to answer the following questions:
%N-CMTF provides a solution to a recommendation or prediction problem where the estimated missing data can be treated as prediction. For example, in LastFM and Delicious datasets that we used in our evaluation, the goal is to predict the missing entries of the tensor as accurately as possible. These entries are then inferred as “most likely items” that can be recommended to users.
% On the other hand, CAN-CSMTF provides a solution to pattern mining.

\textbf{Q1.} What is Cut-CD's runtime performance compared to the state-of-the-art N-CMTF algorithms?

\textbf{Q2.} How accurately Cut-CD can approximate the tensor?

\textbf{Q3.} How accurately Cut-CD can predict the missing data of the tensor? In other words, what is its accuracy performance in a recommendation or prediction task?

\textbf{Q4.} How well Cut-CD with Sparsity Constraint (Cut-CD-SC) can identify factors used in pattern mining?

The datasets used for the experiments, the experimental setups, and the evaluation criteria are detailed in Sections 6.1, 6.2 and 6.3 respectively. Q1, Q2, Q3, and Q4 are addressed in Sections 6.4, 6.5, 6.6, and 6.7 respectively.
     
%-----------
\subsection{Datasets}
Several synthetic and openly available real-world datasets were used in experiments (Table~\ref{table_dataset}). The randomly generated synthetic dataset (Syn1) was generated showing high sparsity. LastFM, Delicious datasets\footnote{https://grouplens.org/datasets/hetrec-2011/} and Syn1 were used to test the tensor approximation performance. The LastFM dataset includes a total of 200 users, 12500 artists and 1719 tags information represented as tensor with a coupled matrix of 200 user’s relationship. The Delicious dataset includes a tensor of 100 users $\times$ 5010 URLs $\times$ 5120 tags with a coupled matrix of 100 user’s relationship. These three datasets are used to address the questions Q1, Q2, and Q3.

We address Q4 by using two smart city data (D1 and D3). The first smart city data (D1) is collected from the activities of Singapore’s elderly people living in the Bukit Panjang region. D1 was gathered using smartphone sensors that consist of 37 users’ home location, 1295 locations visited by them and the time \cite{billy2017extract}. We represent D1 in the tensor model as 37 $\times$ 1295 $\times$ 7 along with user-user physical distance matrix representation 37 $\times$ 37 to identify temporal patterns over 7 days of the week. We changed the time slot in D1 to 24 hours to generate data D2 in the tensor model to identify the temporal pattern over 24 hours. The second smart city data, Tokyo city foursquare\footnote{https://www.kaggle.com/spinalnerve/nyc-tokyo}, consists of 2294 users and their check-ins at 25000 restaurants. 
%This LBSN dataset does not have a distance between two users. 
The dataset was added with an auxiliary information of 385 categories for different locations. We represent ($user \times location \times time$) as a tensor and ($location \times location~category$) as an additional matrix.  We generated two datasets by defining 7 days and 24 hours as the temporal dimension of the tensor as shown in Table~\ref{table_dataset}.

\begin{table}[tb!]
\caption{Dataset statistics.}
\label{table_dataset}
\centering
\begin{tabular}{lllc}
\hline
Dataset & Tensor size & Matrix size & Density of tensor\\
\hline
Syn1   & 1500 x 1500 x 1500 &  1500 x 1500  &  0.00009 \\
LastFM   & 200 x 12500 x 1719 &  200 x 200  &  0.00040 \\
Delicious   & 100 x 5010 x 5120 &  100 x 100  &  0.00070 \\
D1   & 37 x 1295 x 7 &  37 x 37  &  0.00700 \\
D2   & 37 x 1295 x 24 &  37 x 37  &  0.00200 \\
D3   & 25000 x 2294 x 7 &  25000 x 385  &  0.00030 \\
D4   & 25000 x 2294 x 24 &  25000 x 385  &  0.00010 \\
\hline
\end{tabular}
\end{table}

%-------------
\subsection{Experimental setup and Benchmarks} 
The source code of Cut-CD is available\footnote{https://github.com/thirubs/Cut-CD}. All the experiments were executed in Intel (R) Xeon (R) CPU E5-2665 0 @ 2.40GHz model with 16GB RAM.  We compared the performance of Cut-CD with the state-of-the-art N-CMTF factorization algorithms including GCD \cite{balasubramaniam2018nonnegative}, Turbo-SMT OPT \cite{turbo} (an advance variation of OPT \cite{acar2011all}), ALS \cite{hunyadi2014block}, and CCD++ (we implemented the serial version of CCD++ to solve the N-CMTF problem) \cite{ccdpp}. For a fair comparison, we used the single machine implementation of Turbo-SMT OPT by setting parameter repetitions and sample factor to 1. All the algorithms were implemented in MATLAB using the tensor toolbox. Note that CDTF and PCMF are distributed and parallel variations of CCD++ for TF and Coupled Matrix Factorization respectively. Therefore, to have a fair comparison we use the fundamental factorization algorithm CCD++ employed in those methods. %Note that CDTF and PCMF follow the same update sequence as CCD++. Therefore, in this paper we compare the proposed Cut-CD against CCD++. 
CCD++ has a parameter that defines the number of inner iterations it can have. We included two versions of CCD++ (CCD++1 and CCD++3) as benchmarks where CCD++1 stands for CCD++ with 1 inner iteration and CCD++3 stands for CCD++ with 3 inner iterations. Each experiment setting is executed 5 times with different factor matrix initializations and the average value is reported in the result section. 

Identifying the rank of a matrix or a tensor is an NP-hard problem. It is known that the higher the rank, the higher the computation complexity will be, and the solution reaches a stable condition after a certain rank value. In our experiments, we have set the matrix and tensor rank to 100 and 75 for LastFM and Delicious datasets respectively and 30 for Syn1 as the approximation reached stable condition without the further chances of improvement.

\subsection{Evaluation Criteria}
Accuracy is calculated using Normalized Residual Value (NRV) \cite{kimura2015var} as:
\begin{equation}
\label{eq_37}
    \text{Approximation Error (NRV)} = \frac{\norm{\boldsymbol{\mathrm{X_1}} - \boldsymbol{\mathrm{U}}^{(1)}(\boldsymbol{\mathrm{W}} \odot \boldsymbol{\mathrm{V}})^T}^2}{\norm{\boldsymbol{\mathcal{X}}}^2}.
\end{equation}
Prediction Accuracy is reported by the Root Means Square Error (RMSE) and the recommendation quality is evaluated using precision, recall and F1 score, 
\begin{equation}
\label{eq_38}
    \text{RMSE} = \sqrt{\frac{\sum(\boldsymbol{\mathcal{X}}_t - \Hat{\boldsymbol{\mathcal{X}}_t})^2}{n}}
\end{equation}
where \(\boldsymbol{\mathcal{X}}_t\) is the original tensor populated with the test data,  \(\Hat{\boldsymbol{\mathcal{X}}_t}\) is the approximated tensor after the factorization process and $n$ is the number of elements in \(\boldsymbol{\mathcal{X}}_t\).
\begin{equation}
    \label{eq_39}
    %\text{Precision} = \frac{|\{ relevant\_items \}|\cap|\{ retrieved\_items \}| }{|\{ retrieved\_items \}|}.
    \text{Precision} = \frac{True~Positive}{True~Positive + False~Positive}.
\end{equation}
\begin{equation}
    \label{eq_40}
    %\text{Recall} = \frac{|\{ relevant\_items \}|\cap|\{ retrieved\_items \}| }{|\{ relevant\_items \}|}.
    \text{Recall} = \frac{True~Positive}{True~Positive + False~Negative}.
\end{equation}
\begin{equation}
\label{eq_41}
    \text{F1 Score} = 2\left(\frac{Precision \times Recall}{Precision + Recall}\right).
\end{equation}

Pattern Distinctiveness (PD) is used to evaluate the quality of patterns learned using the N-CMTF as follows. 
\begin{equation}
\label{eq_42}
    \text{PD} = <\boldsymbol{\mathrm{t_q}} , \boldsymbol{\mathrm{t_r}}>, \forall q,r \in [1,R], q<r
\end{equation}
where \(\boldsymbol{\mathrm{t_q}}\)  and \(\boldsymbol{\mathrm{t_r}}\)  indicates the \(q^{th}\) and \(r^{th}\) column of temporal factor matrix \(\boldsymbol{\mathrm{W}}\). PD measures the similarity of each pattern with other patterns. Higher the PD value, higher is the similarity between patterns. Since the objective is to identify unique patterns, hence, lower the PD value, better the quality of learned patterns is considered.

%--------------
\subsection{Runtime Performance}
\begin{figure*}
	\centering
	\subfloat[Running time vs Mode length]{{\includegraphics[width=1.8in, height=1.35in]{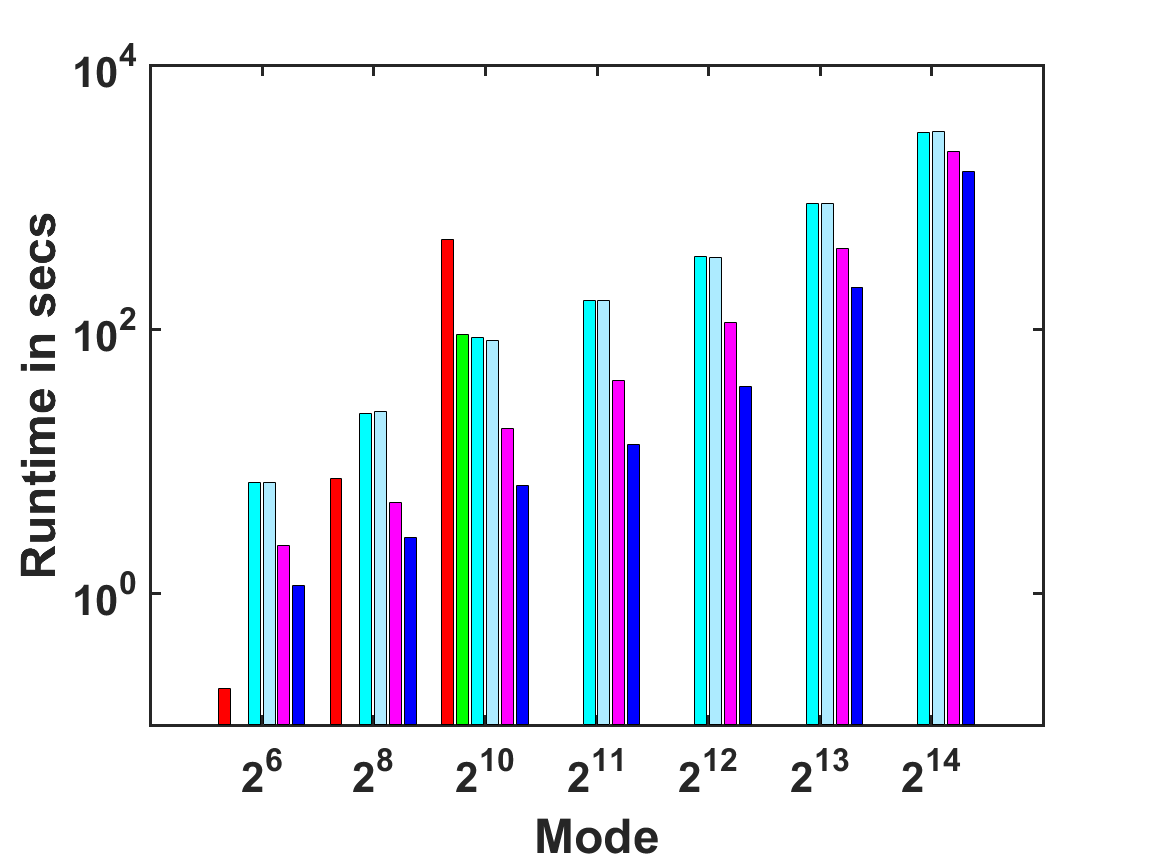}}}%
	%\qquad
	\subfloat[Running time]{{\includegraphics[width=1.8in, height=1.35in]{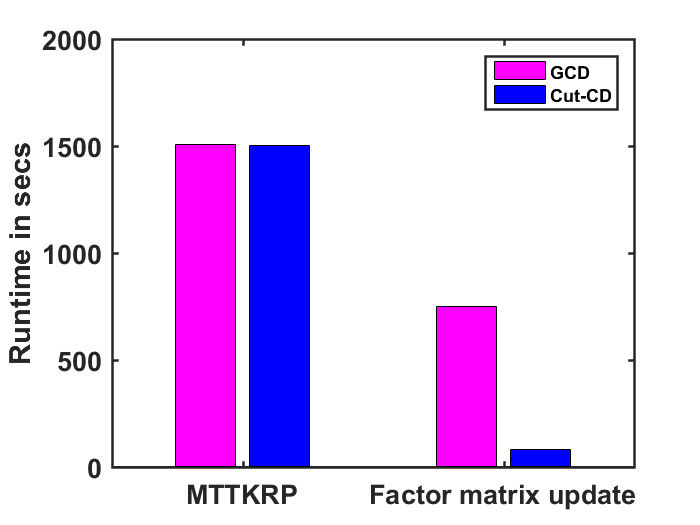}}}%
	%\qquad
	\subfloat[Running time vs Density]{{\includegraphics[width=1.8in, height=1.35in]{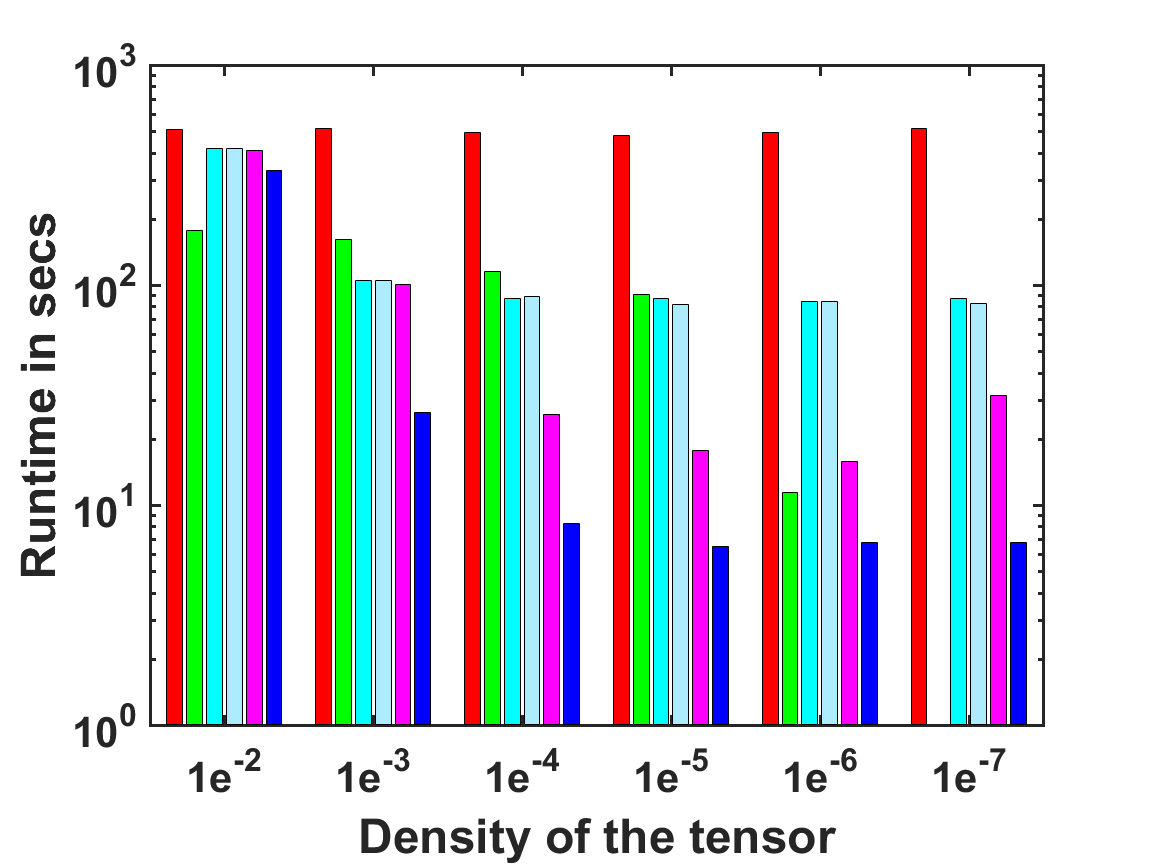}}}%
	%\qquad
	\subfloat[Running time vs Rank]{{\includegraphics[width=1.8in, height=1.35in]{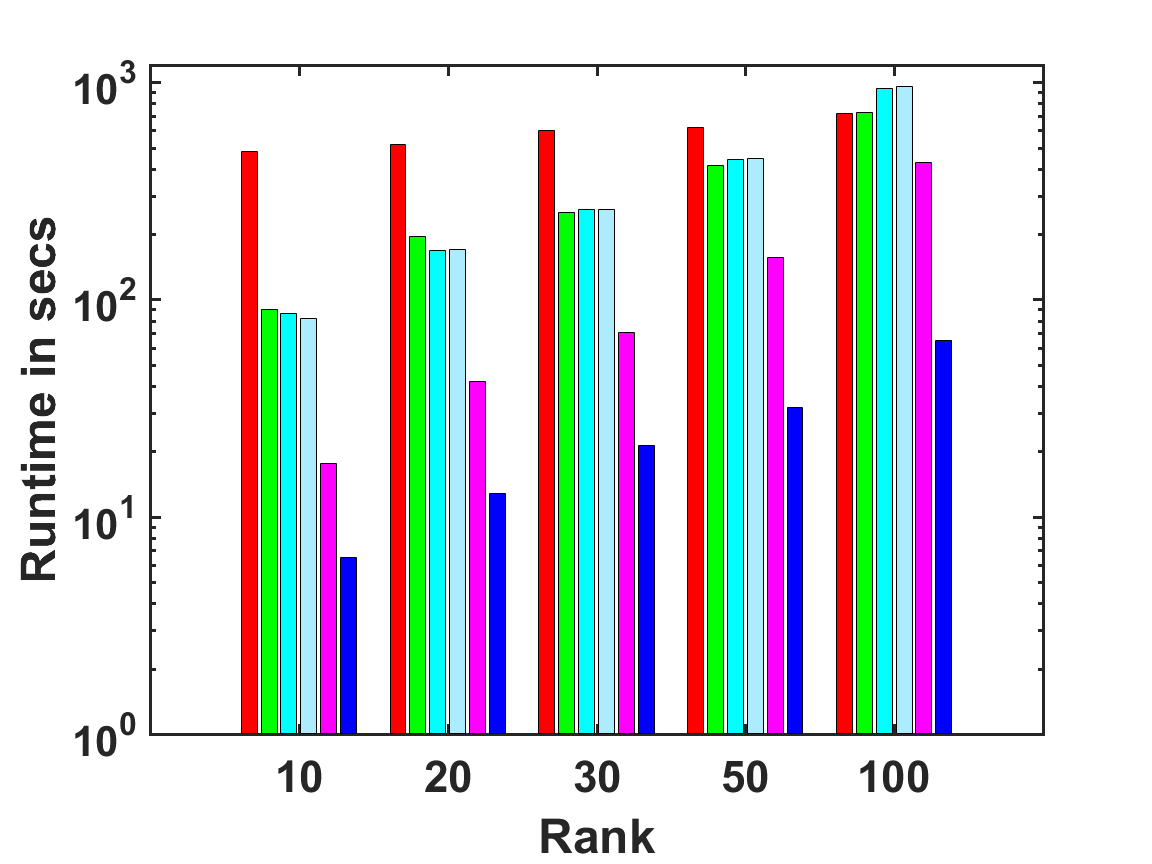}}}%
	\caption{Runtime performance of all algorithms with varying mode length, rank, and density of the tensor. ALS and Turbo-SMT OPT did not scale with mode length $>$ $2^{10}$. Turbo$-$SMT OPT fails to process the tensor for mode length $<$ $2^{10}$ and density $<$ $1e^{-6}$. Legends showing the bar type are given below.}%
	\label{fig:scalability}
\end{figure*}

The runtime of Cut-CD and all benchmarking algorithms in factorizing a tensor has been reported in Fig.~\ref{fig:scalability} by varying the mode length, density and rank of the tensor. We first generate randomly initialized synthetic datasets by increasing the length of mode~\((J,K,L,M)\) of the tensor and the matrix from \(2^6\) to \(2^{14}\), by fixing the rank and the tensor density to \(10\) and \(0.00001\) respectively.  As shown in Fig.~\ref{fig:scalability}a, the runtime of Cut-CD is lower than other algorithms. The runtime of ALS increases exponentially with an increase in data size, whereas Cut-CD, GCD, and CCD++ show the linear growth. Comparing the maximum size of the dataset that ALS and Turbo-SMT OPT can handle, Cut-CD can process a \(2^4\)  times bigger dataset. Most importantly, due to high sparsity, Turbo-SMT OPT even fails to run for smaller datasets (\(2^6\) and \(2^8\)) as there are not enough observations for partitioning the dataset into multiple smaller subsets. 
%\begin{figure*}[!t]
%	\centering
%	\begin{minipage}{.25\textwidth}
%		\centering
%		\includegraphics[width=1.8in, height=1.35in]{images/runtimevsmode.png}
%		%\captionof{figure}{Mode}
%		\label{fig:runtimevsmode}
%	\end{minipage}%
%	\begin{minipage}{.25\textwidth}
%	\centering
%	\includegraphics[width=1.8in, height=1.25in]{images/mttkrpvsfactor.png}
	%\captionof{figure}{Time taken for MTTKRP and factor matrix update.}
%\label{fig:mttkrpvsfactor}
%	\end{minipage}%
%	\begin{minipage}{.25\textwidth}
%		\centering
%		\includegraphics[width=1.8in, height=1.35in]{images/runtimevsdensity.png}
	%	\captionof{figure}{Density of the tensor}
%		\label{fig:runtimevsdensity}
%	\end{minipage}%
%	\begin{minipage}{.25\textwidth}
%	\centering
%	\includegraphics[width=1.8in, height=1.35in]{images/runtimevsrank.png}
	%\captionof{figure}{Rank}
%	\label{fig:runtimevsdensity}
%\end{minipage}%
%\caption{Scalability analysis of all the methods with varying mode length, rank, and density.}
%\label{fig:scalability}
%\end{figure*}
\begin{figure*}
	%\centering
		\vskip -2.0\baselineskip plus -1fil
	\subfloat[RMSE vs Mode length]{{\includegraphics[width=2in, height=1.5in]{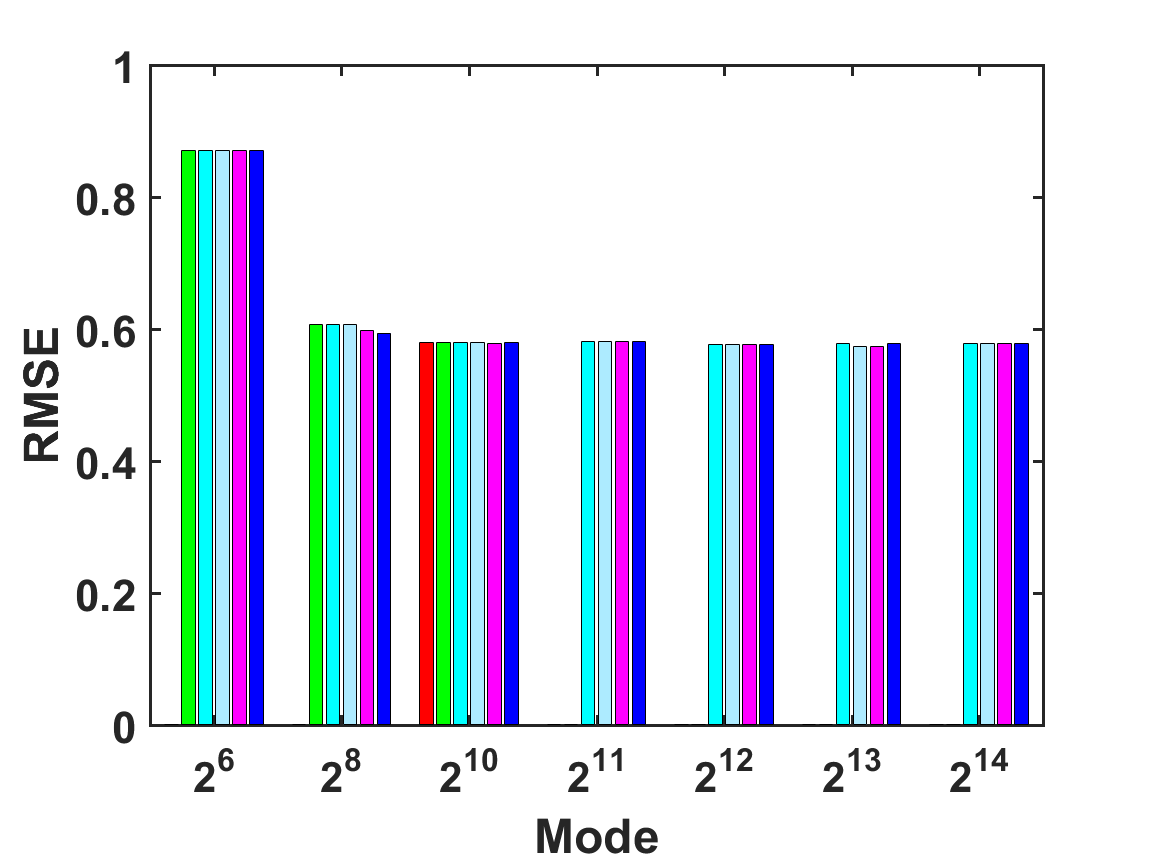}}}%
	%\qquad
	\subfloat[RMSE vs Density]{{\includegraphics[width=2in, height=1.5in]{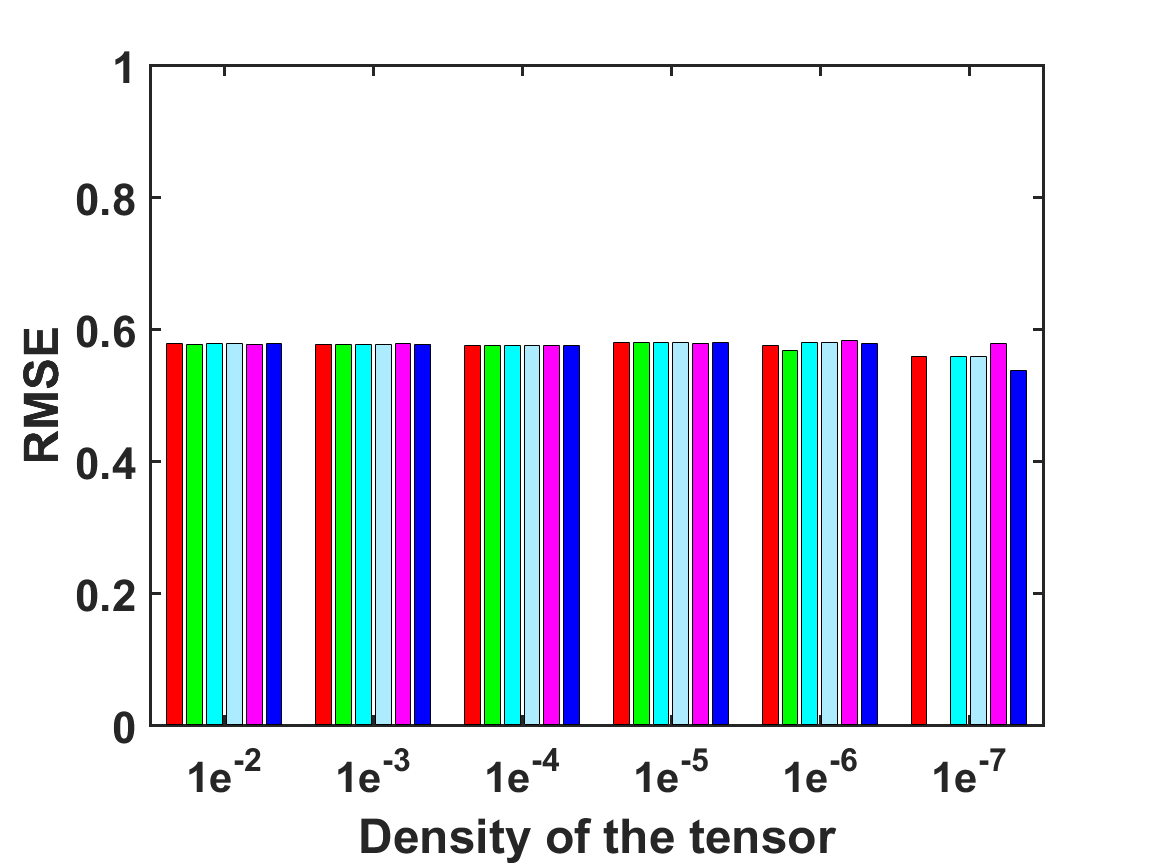}}}%
	%\qquad
	\subfloat[RMSE vs Rank]{{\includegraphics[width=2in,height=1.5in]{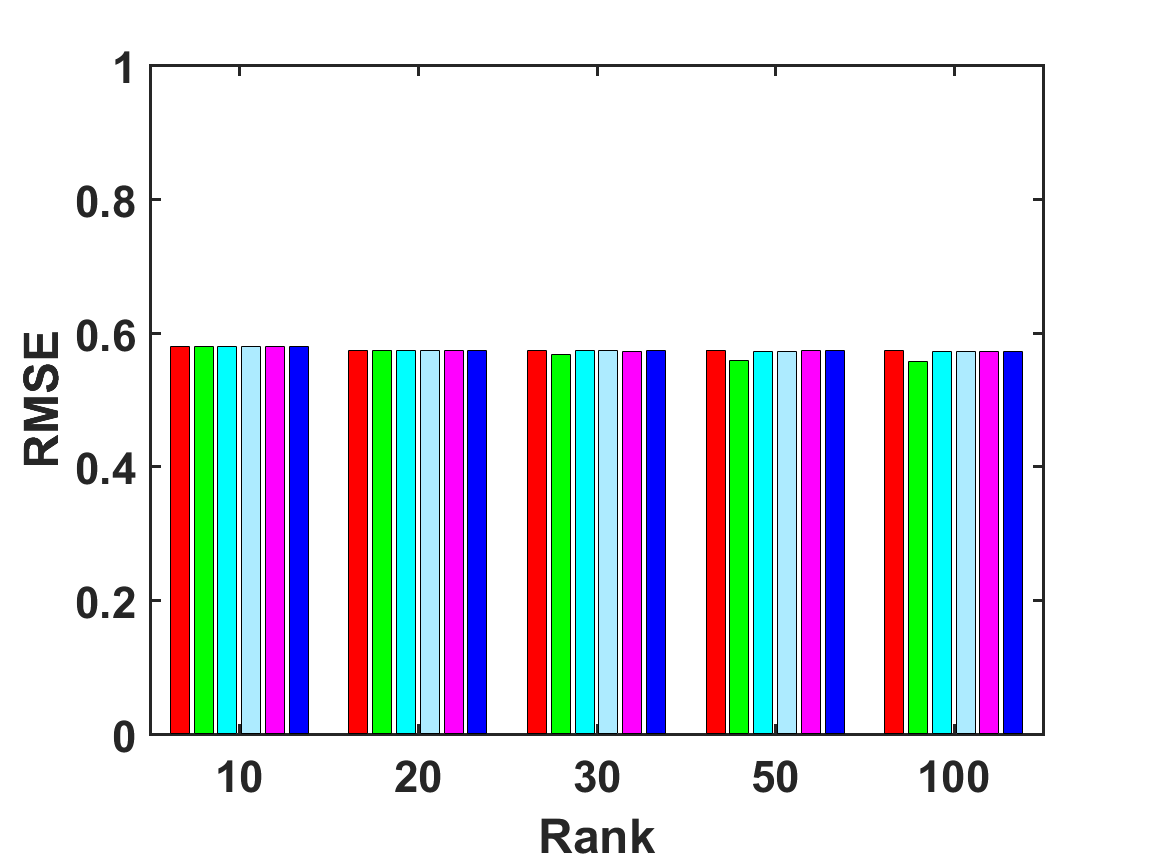}}}\hspace{5mm}%
	\subfloat{\raisebox{7ex}{\includegraphics[width=0.8in, height=0.7in]{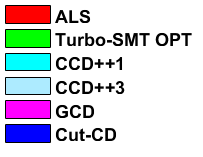}}}
	\caption{Accuracy performance with synthetic datasets (used for scalability evaluation).}%
	\label{fig:scalabilityrmse}
\end{figure*}
%\begin{figure*}[!t]
%	\centering
%	\begin{minipage}{.33\textwidth}
%		\centering
%		\includegraphics[width=1.5in, height=1.25in]{images/rmsevsmode.png}
		%\captionof{figure}{Mode}
%		\label{fig:runtimevsmode}
%	\end{minipage}%
%	\begin{minipage}{.33\textwidth}
%		\centering
%		\includegraphics[width=1.5in, height=1.25in]{images/rmsevsdensity.png}
		%	\captionof{figure}{Density of the tensor}
%		\label{fig:runtimevsdensity}
%	\end{minipage}%
%	\begin{minipage}{.33\textwidth}
%		\centering
%		\includegraphics[width=1.5in, height=1.25in]{images/rmsevsrank.png}
		%\captionof{figure}{Rank}
%		\label{fig:runtimevsdensity}
%	\end{minipage}%
%	\caption{Accuracy performance of all the methods with synthetic datasets used to evaluate the scalability.}
%	\label{fig:scalabilityrmse}
%\end{figure*}

As discussed before, there are two components involved in the time complexity of the factorization process, MTTKRP calculation and factor matrix update. As seen in Fig.~\ref{fig:scalability}b, Cut-CD is nearly 9.2 times faster than GCD in updating the factor matrix while the time taken for MTTKRP calculation remains the same on the \(2^{14}\) mode length dataset. Fig.~\ref{fig:scalability}c shows the runtime on datasets with varying density with the rank and mode length set to 10 and $2^{10}$ respectively. Cut-CD serves the best performance with 4.2 times faster over GCD and 76 times faster over ALS. Performance of Cut-CD increases with increase in the sparsity. For instance, it is 1.5 times faster than ALS for the denser dataset whereas it is 76 times faster for the sparse dataset. This demonstrates the efficiency of Cut-CD for sparse datasets which will be the case for most of the applications. %It shows that Cut-CD serves the best performance with 4.2 times faster over GCD and 76 times faster over ALS. 
In Fig.~\ref{fig:scalability}d, we vary the rank by fixing the density to 0.00001 and mode length to $2^{10}$. We choose this to show the performance of all algorithms, as for larger mode length some of the algorithms (ALS and Turbo-SMT OPT) do not work. Results show that Cut-CD outperforms other algorithms. Even though CCD++ can avoid frequent gradient updates unlike GCD, the runtime performance is inferior to GCD. This is due to the communication cost associated with the frequent change in the factor matrix. It is interesting to note that CCD++1 and CCD++3 show similar performance despite the increase in the inner iterations which shows that the communication cost is the bottleneck and influences the runtime.    %and shows linear performance with the rank. 

%--------------
\subsection{Approximation Performance} 
Fig.~\ref{fig:scalabilityrmse} reports the RMSE performance of all the algorithms for the synthetic datasets used in Section~6.4. It is evident that accuracy is not compromised in Cut-CD for efficient runtime performance. 

\begin{figure*}[ht!]
	\centering
	\subfloat[NRV vs Running time]{{\includegraphics[width=1.8in, height=1.35in]{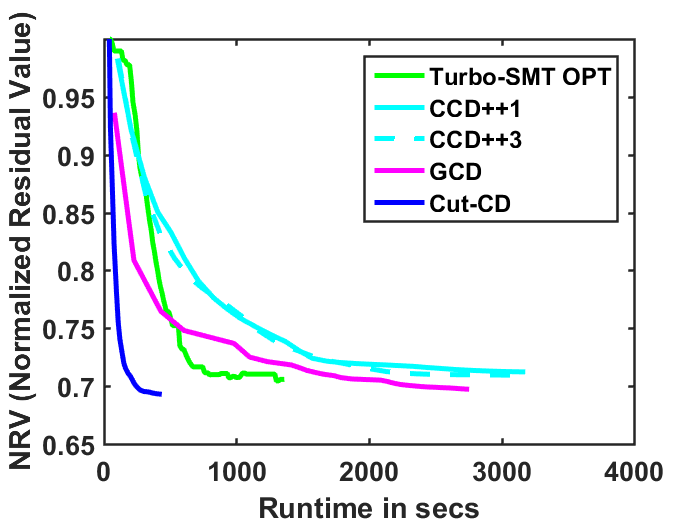}}}%
	%\qquad
	\subfloat[NRV vs Iterations]{{\includegraphics[width=1.8in, height=1.35in]{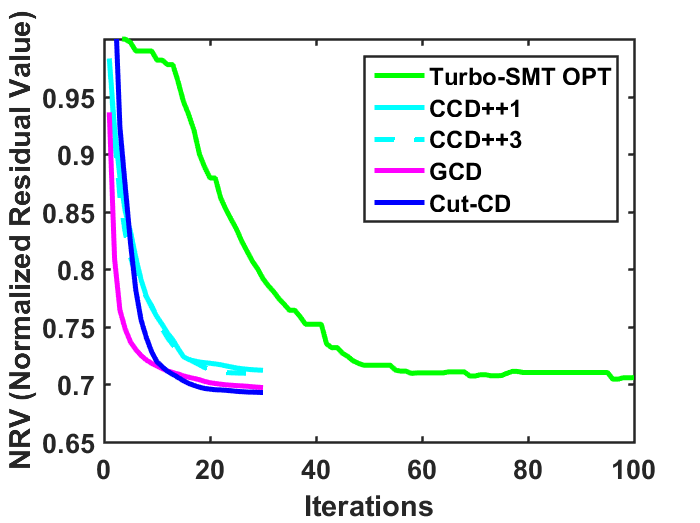}}}%
	%\qquad
	\subfloat[Running time vs Iteration]{{\includegraphics[width=1.8in, height=1.35in]{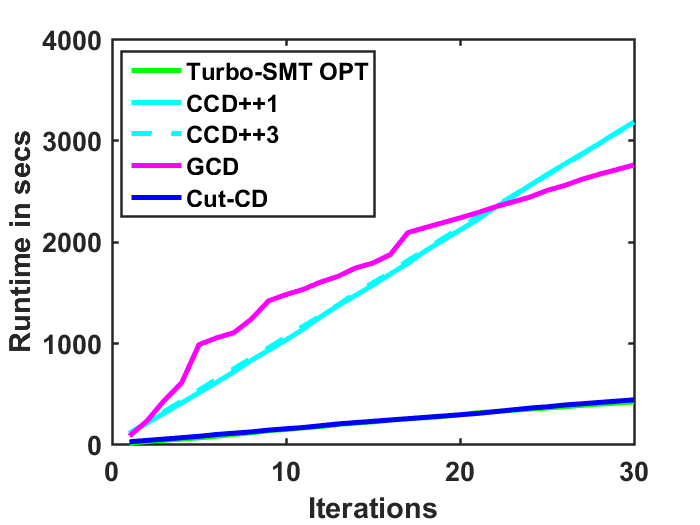}}}%
	%\qquad
	\subfloat[Threshold sensitivity]{{\includegraphics[width=1.8in, height=1.35in]{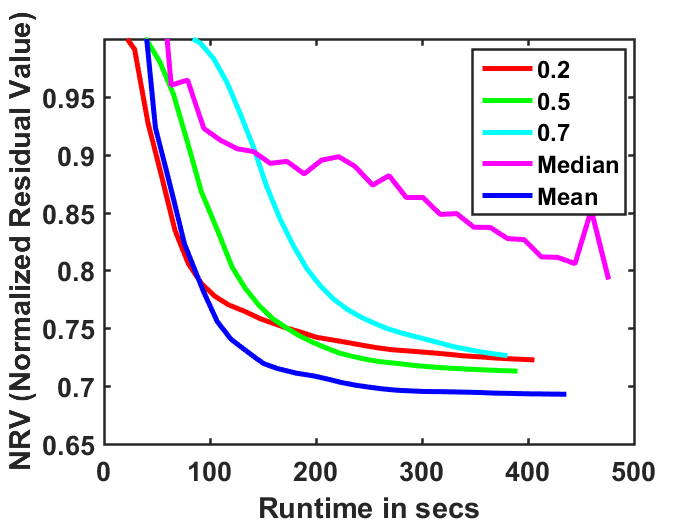}}}%
	\caption{Approximation and Runtime performance of all the algorithms for LastFM dataset.}%
	\label{fig:lastfm}
\end{figure*}

\begin{figure*}[ht!]
	\centering
	\subfloat[NRV vs Running time]{{\includegraphics[width=1.8in, height=1.35in]{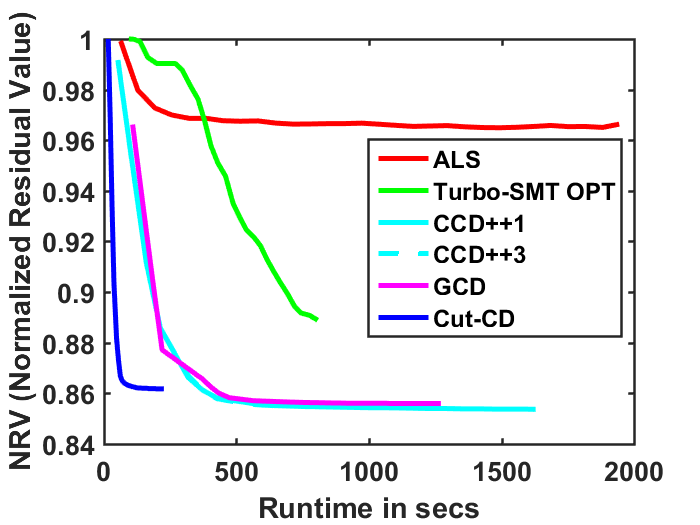}}}%
	%\qquad
	\subfloat[NRV vs Iterations]{{\includegraphics[width=1.8in, height=1.35in]{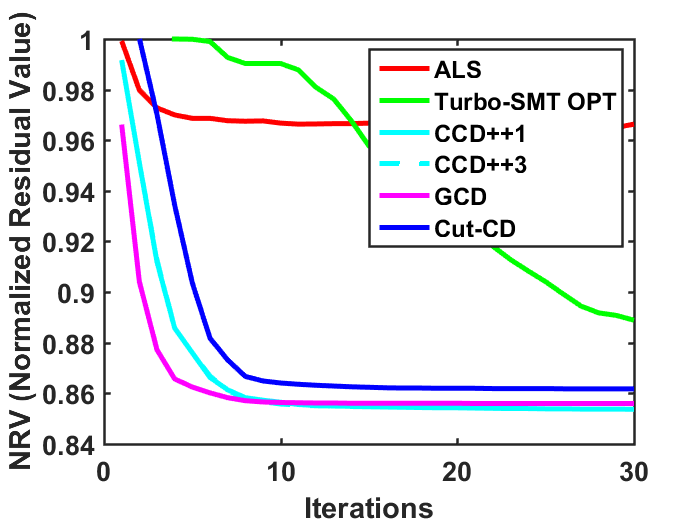}}}%
	%\qquad
	\subfloat[Running time vs Iteration]{{\includegraphics[width=1.8in, height=1.35in]{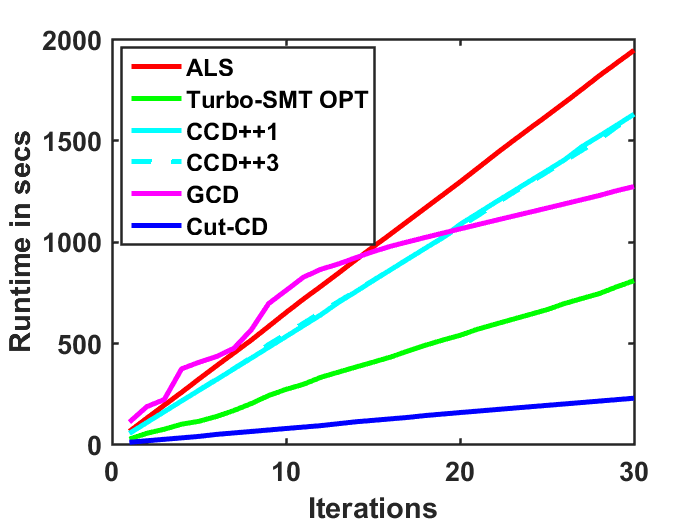}}}%
	%\qquad
	\subfloat[Threshold sensitivity]{{\includegraphics[width=1.8in, height=1.35in]{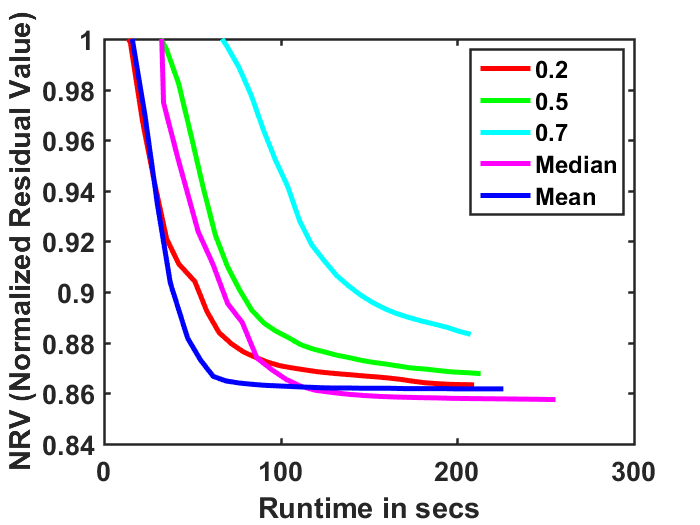}}}%
	\caption{Approximation and Runtime performance of all the algorithms for Delicious dataset.}%
	\label{fig:delicious}
\end{figure*}

\begin{figure*}[ht!]
	\centering
	\subfloat[NRV vs Running time]{{\includegraphics[width=1.8in, height=1.35in]{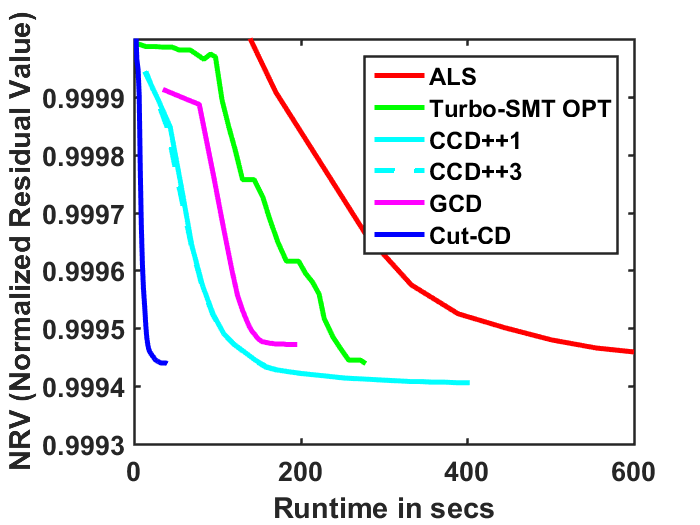}}}%
	%\qquad
	\subfloat[NRV vs Iterations]{{\includegraphics[width=1.8in, height=1.35in]{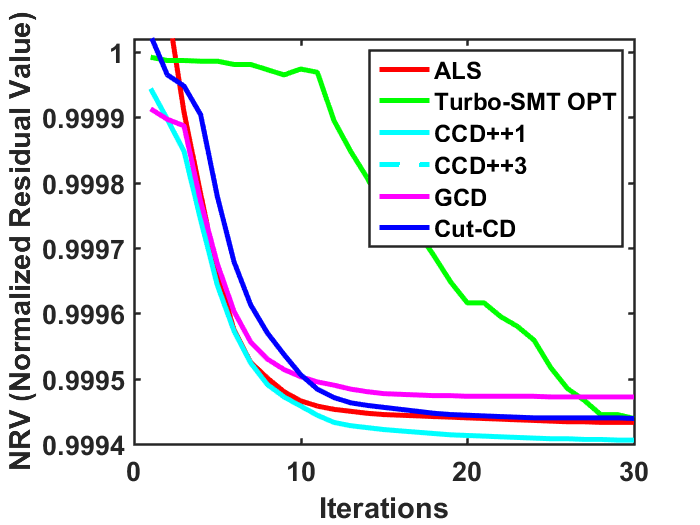}}}%
	%\qquad
	\subfloat[Running time vs Iteration]{{\includegraphics[width=1.8in, height=1.35in]{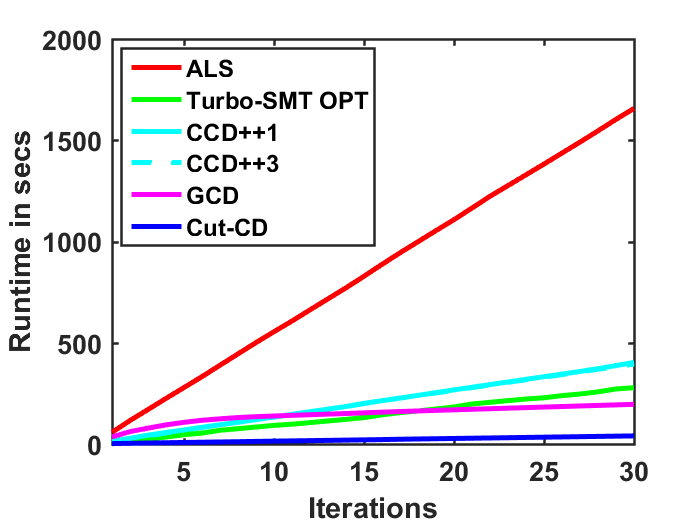}}}%
	%\qquad
	\subfloat[Threshold sensitivity]{{\includegraphics[width=1.8in, height=1.35in]{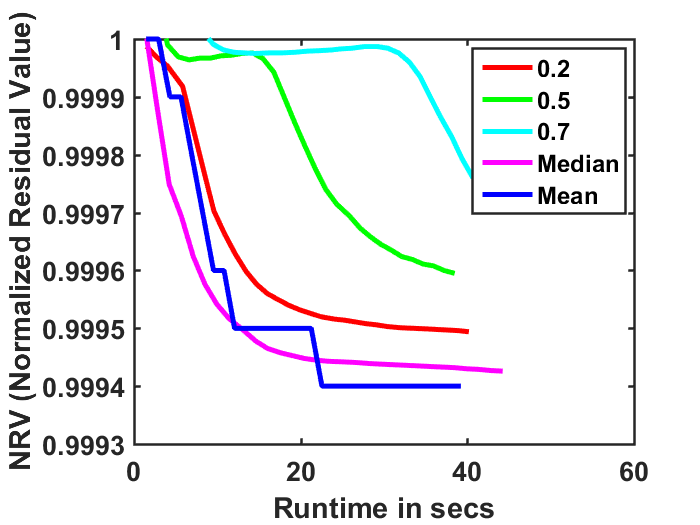}}}%
	\caption{Approximation and Runtime performance of all the algorithms for Syn1 dataset.}%
	\label{fig:syn1}
\end{figure*}

Figures~\ref{fig:lastfm}, \ref{fig:delicious}, and \ref{fig:syn1} show the comparative performance of all algorithms in terms of accuracy and convergence. In all datasets, Cut-CD and GCD converged into a better solution faster without compromising accuracy. For the sparse data (Syn1) in Fig.~\ref{fig:syn1}, Cut-CD converged to a better solution before GCD and ALS algorithms, completing only a single iteration. This indicates the effectiveness of Cut-CD in sparse conditions which is a common situation in real-world applications where other algorithms may suffer.  

%----%%%%
\textit{Sensitivity to cut-off value:} 
Cut-CD requires a cut-off value to select the elements for updating during factorization. By normalizing each column values, a cut-off value can be set between 0 and 1. Experiments with all datasets including the very sparse Syn1 reveal that the $mean$ value performs consistently best (Figs.~\ref{fig:lastfm}d, \ref{fig:delicious}d, and \ref{fig:syn1}d), as well as, it takes least the time to converge. Therefore, the cut-off value is set to $mean$ that makes Cut-CD a parameter-free method. As shown in Fig.~\ref{fig:delicious}a, the accuracy of Cut-CD with mean as cut-off value is less than that of GCD for the Delicious dataset. But with a proper cut-off value selection, its accuracy can further be improved. This is supported by Fig.~\ref{fig:delicious}d.

%\begin{figure*}[ht!]
%    \centering
%    \subfloat[LastFM]{{\includegraphics[width=5.2cm,height=4.5cm]{fig7a.png} }}%
%    \qquad
%    \subfloat[Delicious]{{\includegraphics[width=5.2cm,height=4.5cm]{fig7b.png} }}%
%     \qquad
 %   \subfloat[Syn1]{{\includegraphics[width=5.2cm,height=4.5cm]{fig7c.png} }}%
%    \caption{Runtime Performance}%
%    \label{figruntime}
%\end{figure*}

%\begin{figure*}[ht]
%    \centering
%    \subfloat[LastFM]{{\includegraphics[width=5.2cm,height=4.5cm]{fig8a.png} }}%
%    \qquad
%    \subfloat[Delicious]{{\includegraphics[width=5.2cm,height=4.5cm]{fig8b.png} }}%
%     \qquad
%    \subfloat[Syn1]{{\includegraphics[width=5.2cm,height=4.5cm]{fig8c.png} }}%
%    \caption{Approximation Performance (NRV vs Runtime)}%
%    \label{fignrv}
%\end{figure*}

%\begin{figure*}[ht]
%    \centering
%    \subfloat[LastFM]{{\includegraphics[width=5.2cm,height=4.5cm]{fig9a.png} }}%
%    \qquad
%    \subfloat[Delicious]{{\includegraphics[width=5.2cm,height=4.5cm]{fig9b.png} }}%
%     \qquad
%    \subfloat[Syn1]{{\includegraphics[width=5.2cm,height=4.5cm]{fig9c.png} }}%
%    \caption{Influence of different Cut-off values (NRV vs Iterations)}%
%    \label{figinflu}
%\end{figure*}

%---------------
\subsection{Prediction or Recommender Performance} 
The N-CMTF can be considered as a solution to a recommendation or prediction problem, where the estimated missing data are treated as the prediction. The factor matrices learned during factorization using Cut-CD are used to reconstruct the approximated tensor as per \eqref{eq_5} that will identify missing values. The goal of the recommendation task conducted with LastFM and Delicious datasets is to predict the missing entries of the tensor as accurately as possible. These entries are then inferred as “most likely items” that can be recommended to users.

\begin{figure*}[ht!]
    \centering
    \subfloat[RMSE vs Rank]{{\includegraphics[width=1.8in, height=1.35in]{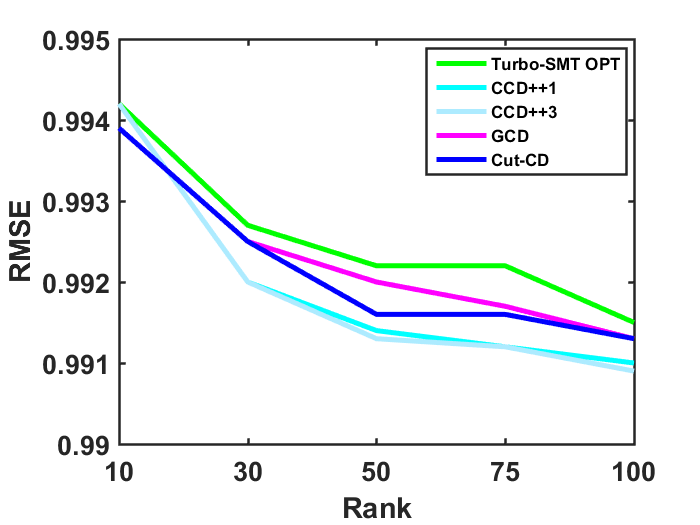}}}%
    %\qquad
    \subfloat[Running time vs Rank]{{\includegraphics[width=1.8in, height=1.35in]{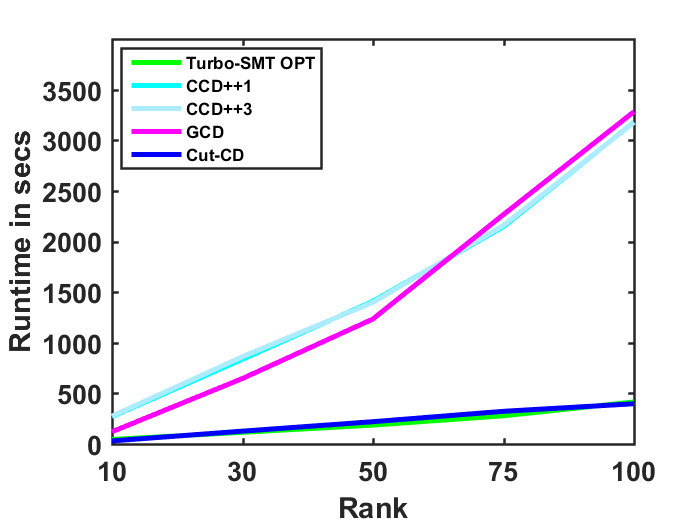}}}%
    %\qquad
   % \label{fig:example}%
    \subfloat[Precision and Recall]{{\includegraphics[width=1.8in, height=1.35in]{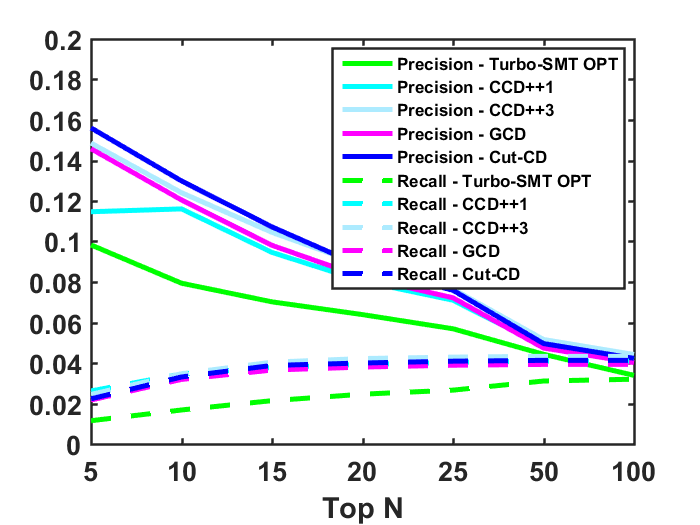}}}%
    %\qquad
    \subfloat[F1 score]{{\includegraphics[width=1.8in, height=1.35in]{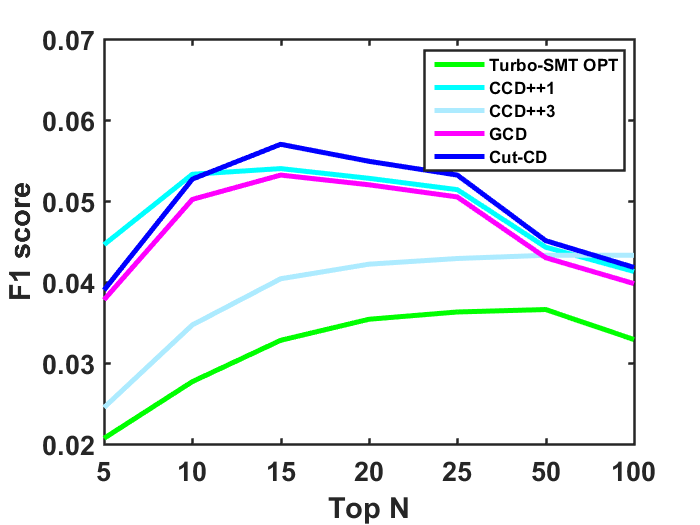}}}%
 \caption{RMSE, Runtime, Precision, Recall and F1 score on the LastFM dataset.}%
 \label{figrmselastfm}%
 \end{figure*}
 
\begin{figure*}[ht!]
	\centering
	\subfloat[RMSE vs Rank]{{\includegraphics[width=1.8in, height=1.35in]{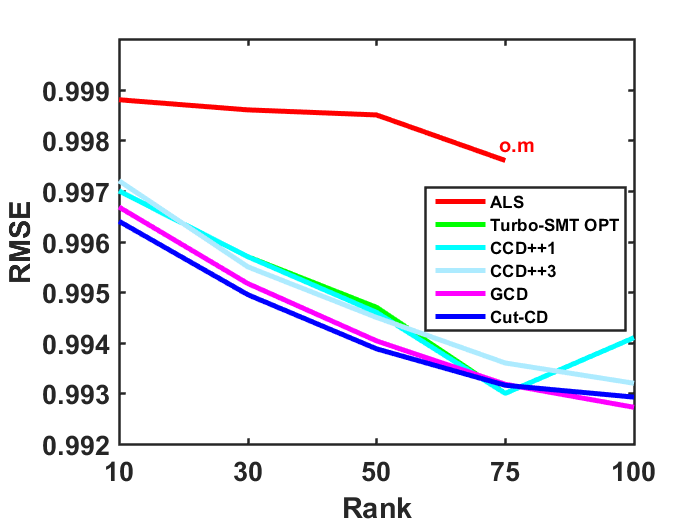}}}%
	%\qquad
	\subfloat[Running time vs Rank]{{\includegraphics[width=1.8in, height=1.35in]{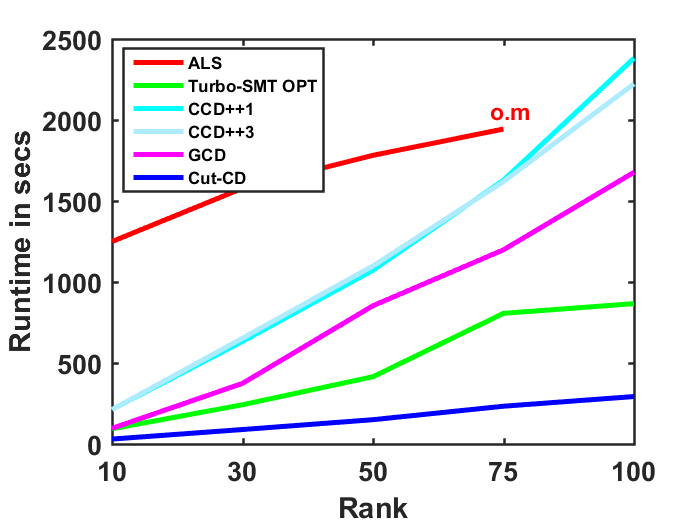}}}%
	%\qquad
	% \label{fig:example}%
	\subfloat[Precision and Recall]{{\includegraphics[width=1.8in, height=1.35in]{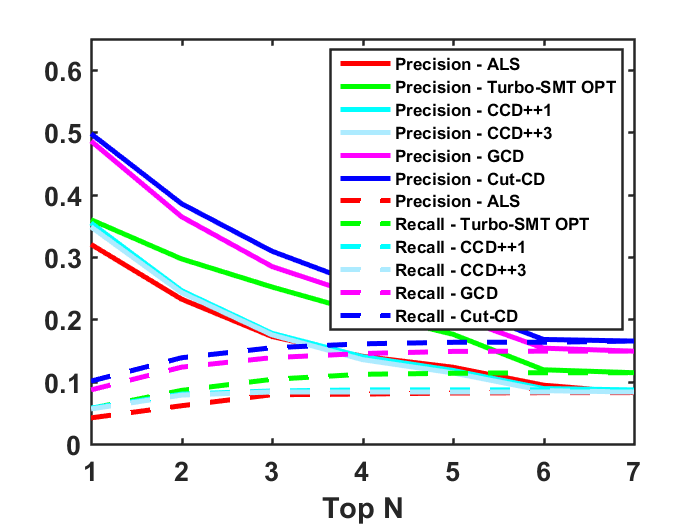}}}%
	%\qquad
	\subfloat[F1 score]{{\includegraphics[width=1.8in, height=1.35in]{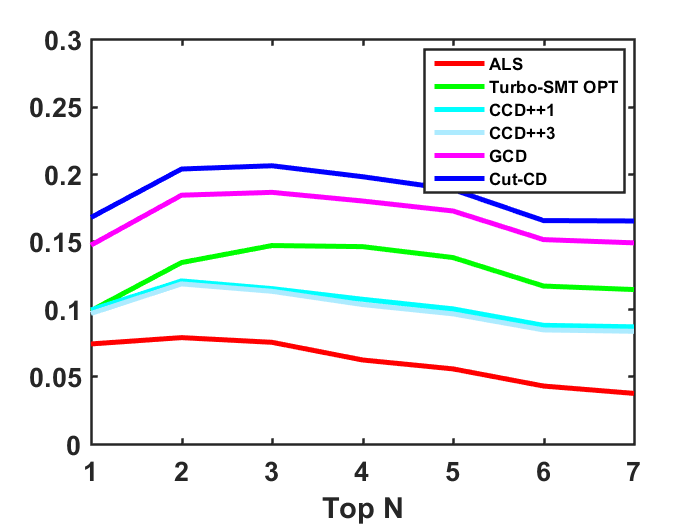}}}%
	\caption{RMSE, Runtime, Precision, Recall and F1 score on the Delicious dataset. o.m. out of memory.}%
	\label{figrmsedel}%
\end{figure*}

Figs.~\ref{figrmselastfm} and \ref{figrmsedel} report the best performance of Cut-CD in comparison to all baseline algorithms. While other algorithms including Cut-CD can handle higher ranks, ALS is not scalable to LastFM and ran out of memory for rank greater than 75 for the Delicious dataset. With lower ranks, it has failed in recommendation generation yielding the lowest F1 score. Though Turbo-SMT OPT can handle higher ranks, due to random sampling it has lost some information leading to poor performance in terms of prediction and recommendation generation. GCD and CCD++ show poorer accuracy performance in comparison to Cut-CD, as well as, GCD is 5.7 to 8.3 times slower and CCD++ is 8 times slower than Cut-CD.

%ALS is not included in the analysis of LastFM dataset as it cannot handle the dataset of this size. Similar to Delicious dataset, Cut-CD shows better performance in LastFM dataset as shown in Fig.~\ref{figrmselast}. The runtime of GCD grows exponentially in this dataset and Cut-CD is 8.3 times faster than GCD. On the other hand, Turbo-SMT ran out of memory for the rank higher than 75 and results in poor recommendation generation.

%Experimental results confirm that Cut-CD is fast and efficient in missing value prediction (i.e. recommendation generation) and can handle larger datasets and higher ranks.

%----------------
\vskip -5\baselineskip plus -1fil
\subsection{Pattern Mining}
The objection function \eqref{eq_8} is designed to generate denser factor matrices so that the reconstructed tensor can be approximated to the original tensor as close as possible. The new or unobserved entries in the approximated tensor can be used for making prediction or recommendation. However, for the purpose of pattern mining, it is better to generate sparse and distinct factor matrices \cite{balasubramaniam2018understanding}. We introduce an auxiliary term to the N-CMTF objective function \eqref{eq_8} that controls the sparsity level of the factor matrices. We add the \(L_{2,1}\) norm to the objective function \eqref{eq_8} as,
\begin{multline}
\label{eq_29}
\min_{\boldsymbol{\mathrm{U}}^{(1)},\boldsymbol{\mathrm{V}}, \boldsymbol{\mathrm{W}}, \boldsymbol{\mathrm{U}}^{(2)} \geq 0} f(\boldsymbol{\mathrm{U}}^{(1)},\boldsymbol{\mathrm{V}}, \boldsymbol{\mathrm{W}}, \boldsymbol{\mathrm{U}}^{(2)}) = \\
\norm{\boldsymbol{\mathcal{X}} - \llbracket\boldsymbol{\mathrm{U}}^{(1)}, \boldsymbol{\mathrm{V}}, \boldsymbol{\mathrm{W}} \rrbracket }^2 + \left\lVert\boldsymbol{\mathrm{Y}} - \boldsymbol{\mathrm{U}}^{(1)} \boldsymbol{\mathrm{U}}^{(2)}\right\rVert^2 +\lambda \norm{\boldsymbol{\mathrm{U}}^{(1)}}_{2,1} \\
+\lambda \norm{\boldsymbol{\mathrm{V}}}_{2,1}
+\lambda \norm{\boldsymbol{\mathrm{W}}}_{2,1}+\lambda \norm{\boldsymbol{\mathrm{U}}^{(2)}}_{2,1}
\end{multline}
where \(\norm{\boldsymbol{\mathrm{U}}^{(1)}}_{2,1}\), \(\norm{\boldsymbol{\mathrm{V}}}_{2,1}\), \(\norm{\boldsymbol{\mathrm{W}}}_{2,1}\)  and \(\norm{\boldsymbol{\mathrm{U}}^{(2)}}_{2,1}\) are \(L_{2,1}\) norms applying the sparsity constraints on each factor matrices and $\lambda$ indicates the regularization parameter.

The $L_{2,1}$ norm of factor matrices say for \(\norm{\boldsymbol{\mathrm{U}}^{(1)}}_{2,1}\) can be calculated as,
\begin{equation}
\label{eq_30}
\norm{\boldsymbol{\mathrm{U}}^{(1)}}_{2,1} = \sum_{j = 1}^{J}\sqrt{\sum_{r = 1}^{R}(\mathrm{u}^{(1)}_{jr})^2} = \sum_{j = 1}^{J}\norm{\boldsymbol{\mathrm{u^{(1)}_{*j}}}}. 
\end{equation}

\begin{comment}
The objective function formulated in \eqref{eq_29} is a non-quadratic and non-convex, hence it will be difficult to solve. The half-quadratic minimization technique will convert the non-quadratic function in to an augmented objective function that eases the minimization process~\cite{du2012robust}. Hence, we propose to calculate $L_{2,1}$ norm of factor matrices say for \(\norm{\boldsymbol{\mathrm{U}}^{(1)}}_{2,1}\) as using an auxiliary variable based on half-quadratic minimization as,
%Using the half-quadratic minimization method~\cite{du2012robust}, we can calculate \(L_{2,1}\) norm, say for \(\norm{\boldsymbol{\mathrm{U}}^{(1)}}_{2,1}\) as
%
\begin{equation}
\label{eq_30}
\norm{\boldsymbol{\mathrm{U}}^{(1)}}_{2,1} = {Tr}(\boldsymbol{\mathrm{U}}^{(2)^T} \boldsymbol{\mathrm{Q}}_{\boldsymbol{\mathrm{U}}^{(1)}}\boldsymbol{\mathrm{U}}^{(1)})
\end{equation}
where \(\boldsymbol{\mathrm{Q}}_{\boldsymbol{\mathrm{U}}^{(1)}}\) is a diagonal matrix whose \(j^{th}\)  diagonal elements are computed as \(q_{u^{(1)}_{jj}}  =  \frac{1}{2\norm{\boldsymbol{\mathrm{u}}^{(1)}_j}_1}\) . \(\boldsymbol{\mathrm{u}}^{(1)}_j\) indicates the \(j^{th}\) row of \(\boldsymbol{\mathrm{U}}^{(1)}\). We calculate \(\boldsymbol{\mathrm{Q}}_{\boldsymbol{\mathrm{V}}}\) , \(\boldsymbol{\mathrm{Q}}_{\boldsymbol{\mathrm{W}}}\)  and \(\boldsymbol{\mathrm{Q}}_{\boldsymbol{\mathrm{U}}^{(2)}}\) similarly for other factor matrices \(\boldsymbol{\mathrm{V}}\), \(\boldsymbol{\mathrm{W}}\) and \(\boldsymbol{\mathrm{U}}^{(2)}\) respectively.
\end{comment}
With the sparsity constraint added to the objective function, the non-zero elements in the factor matrix will be reduced leading to a sparse factor matrix. As the sparsity constraint in \eqref{eq_29} is introduced using norm, it will not change the optimization process and we can easily apply Cut-CD for N-CMTF with Sparsity Constraint (Cut-CD-SC) by repeating the Sections 4.2 to 4.5 and calculating \(L_{2,1}\)  of each factor matrix using \eqref{eq_30}.

\subsubsection{Understanding the Spatio-Temporal Patterns of Singapore’s Elderly people (D1 and D2)}
In this section, we apply Cut-CD-SC to extract spatio-temporal patterns from the smart city datasets, D1 and D2. We represent the users' mobility information in the N-CMTF of ($user \times location \times time~slots$) with ($user \times user$) to identify spatio-temporal patterns over 7 days of the week and over 24 hours in D1 and D2 respectively. Though NMF can be useful to extract spatial-temporal patterns separately by representing ($user \times location$) and ($user \times time~slots$) matrices, they cannot be used to capture spatial and temporal patterns existing together. %Therefore, N-CMTF is an obvious choice to model this multi-dimensional data and automate the process of spatio-temporal pattern extraction.
With rank ($R$) set as 4 during the factorization process, 4 distinct patterns relating to each of the temporal ($\boldsymbol{\mathrm{W}} \in \R^{L \times R}$) and location ($\boldsymbol{\mathrm{V}}\in \R^{K \times R}$) factor matrices are obtained. Each column in the factor matrix becomes a pattern/feature. 

\begin{figure*}[ht!]
	\centering
	%\subfloat[Time factors on D1]{{\includegraphics[width=1.8in, height=1.35in]{images/d1_temporal.png}}}%
	\subfloat[Time factors on D1]{{\includegraphics[width=1.8in, height=1.35in]{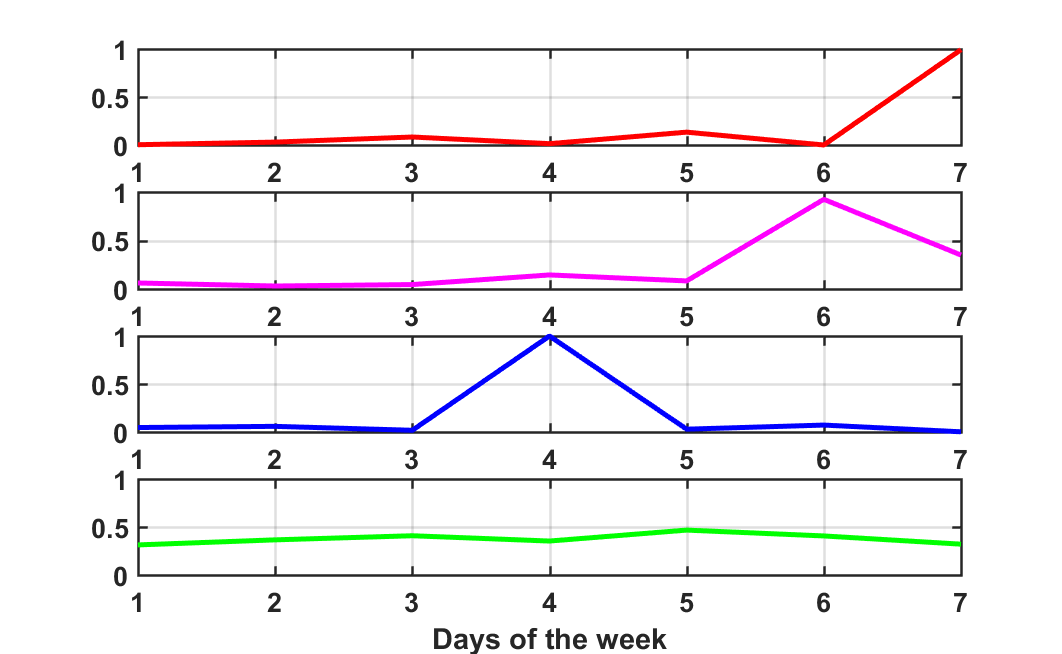}}}%
	%\qquad
	%\subfloat[Time factors on D2]{{\includegraphics[width=1.8in, height=1.35in]{images/d2_temporal.png}}}%
	\subfloat[Time factors on D2]{{\includegraphics[width=1.8in, height=1.35in]{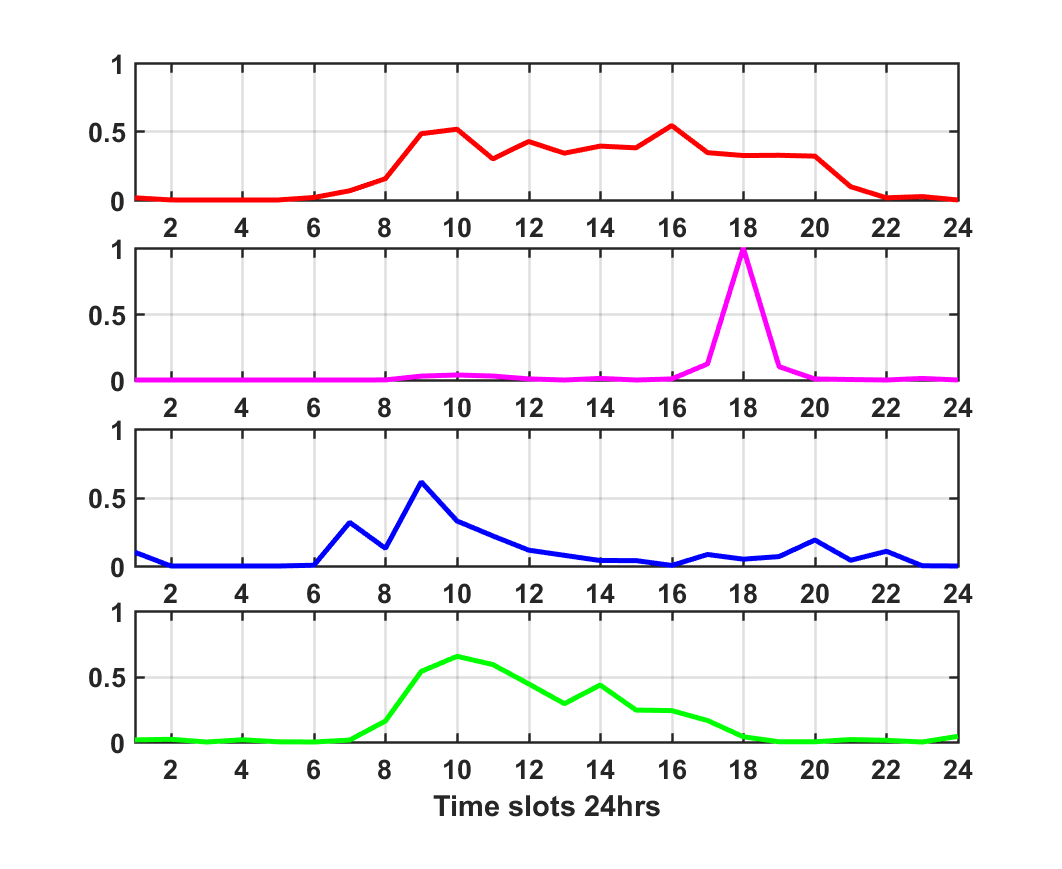}}}%
	%\qquad
	%\subfloat[Time factors on D3]{{\includegraphics[width=1.8in, height=1.35in]{images/d3_temporal.png}}}%
	\subfloat[Time factors on D3]{{\includegraphics[width=1.8in, height=1.35in]{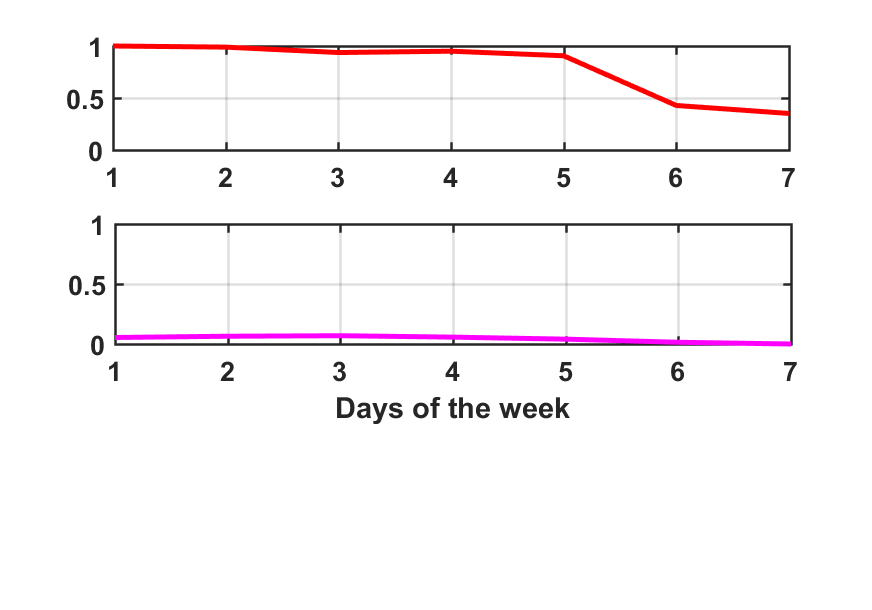}}}%
	%\qquad
	%\subfloat[Time factors on D4]{{\includegraphics[width=1.8in, height=1.35in]{images/d4_temporal.png}}}%
	\subfloat[Time factors on D4]{{\includegraphics[width=1.8in, height=1.35in]{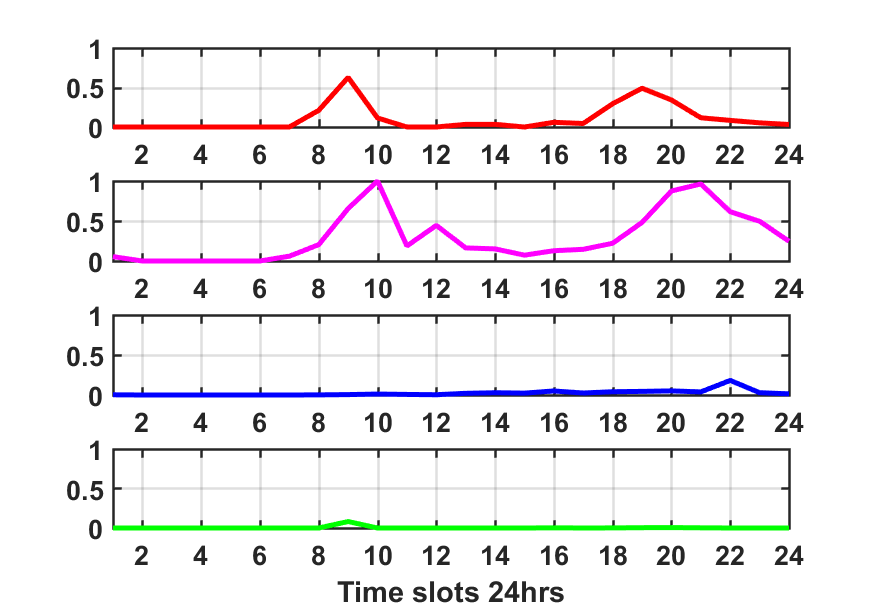}}}%
	\caption{Temporal patterns derived from the 3rd mode (time) of the tensor using Cut-CD-SC. $y-axis$ shows the normalized values of elements in a column of the factor matrix.}%
	\label{fig:temporal}
\end{figure*}

Fig.~\ref{fig:temporal}a and Fig.~\ref{fig:temporal}b show the temporal patterns generated by Cut-CD-SC on D1 and D2 where different color indicates different patterns. The red and pink patterns in Fig.~\ref{fig:temporal}a show that people are more active on weekends (Sundays and Saturdays respectively) and stay inactive over weekdays. The blue pattern indicates that some users are active on Thursdays and remain inactive during the weekend. The red and pink patterns in Fig.~\ref{fig:temporal}b provide fine-granular information that people are engaged in activities between 6 - 10 pm, and in the evening at 6 pm, respectively. The blue pattern in Fig.~\ref{fig:temporal}b shows a peak at 6 to 9 am to distinguish itself from green pattern. %This blue pattern distinctly shows that users are moderately active between 7 am to 1 pm and some are active at 6 to 8 pm. 

\begin{figure*}[ht!]
	\centering
	\subfloat[Spatial Pattern on D1]{{\includegraphics[width=1.65in, height=1.3in]{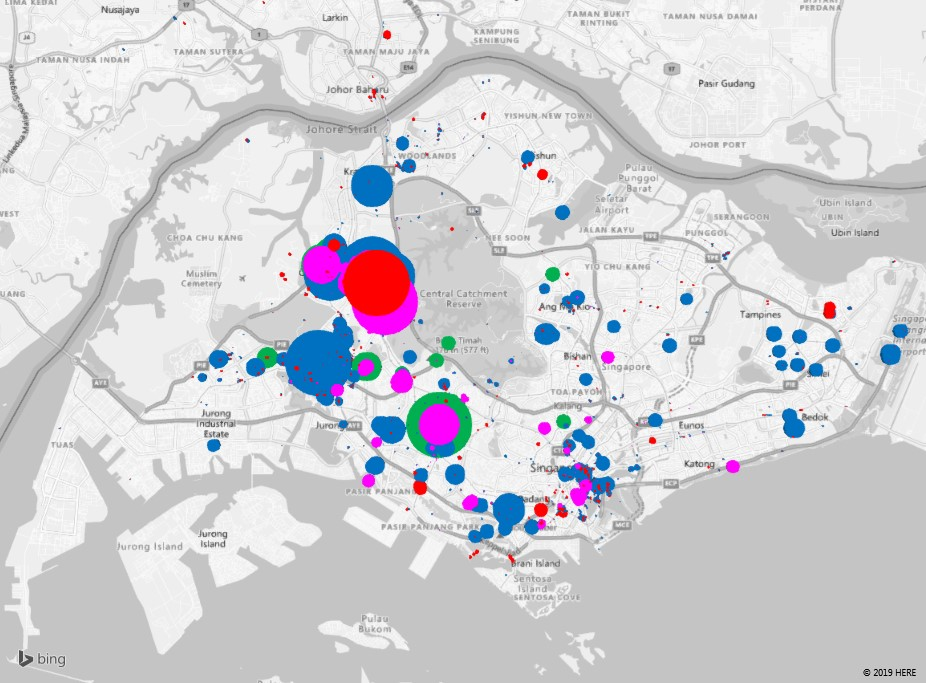}}}%
	%\qquad
	\hspace*{5px}
	\subfloat[Spatial Pattern on D2]{{\includegraphics[width=1.65in, height=1.3in]{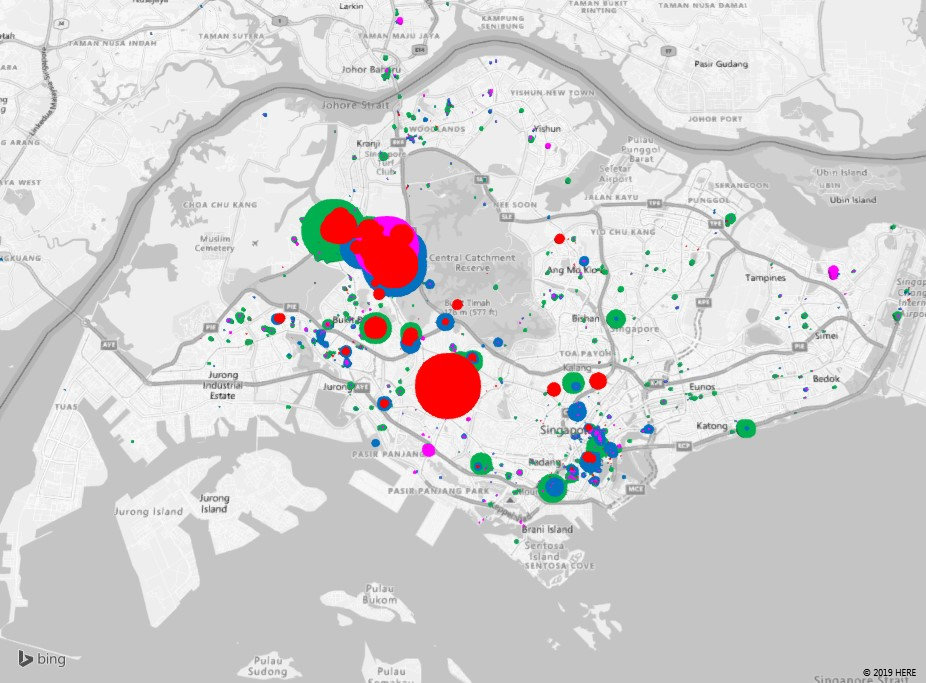}}}%
	%\qquadhttps://www.overleaf.com/3411893283gsqhhvfzjjng
	\hspace*{5px}
	\subfloat[Spatial Pattern on D3]{{\includegraphics[width=1.65in, height=1.3in]{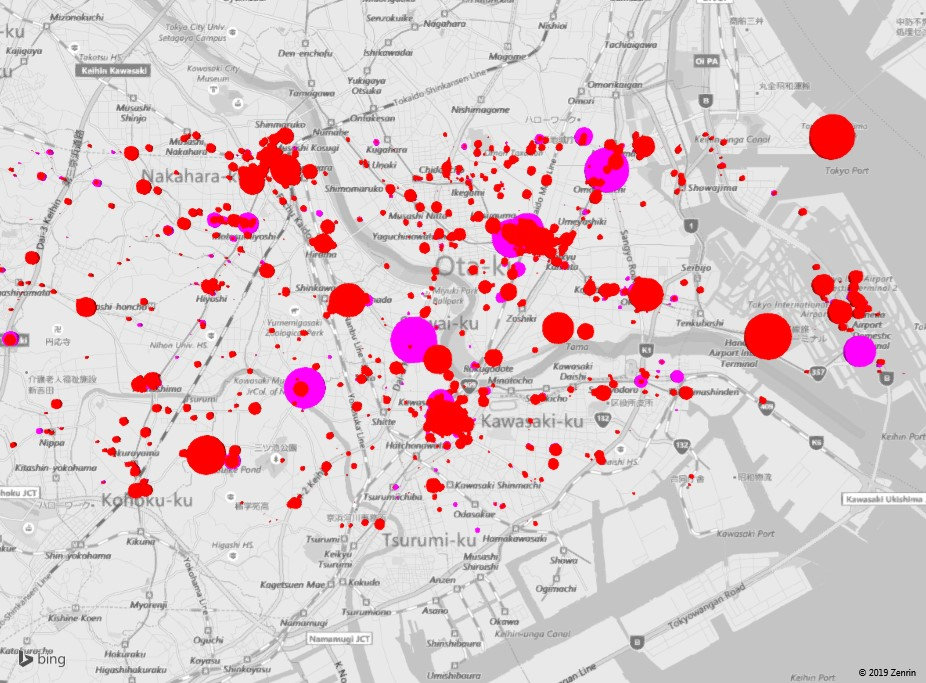}}}%
	%\qquad
	\hspace*{5px}
	\subfloat[Spatial Pattern on D4]{{\includegraphics[width=1.65in, height=1.3in]{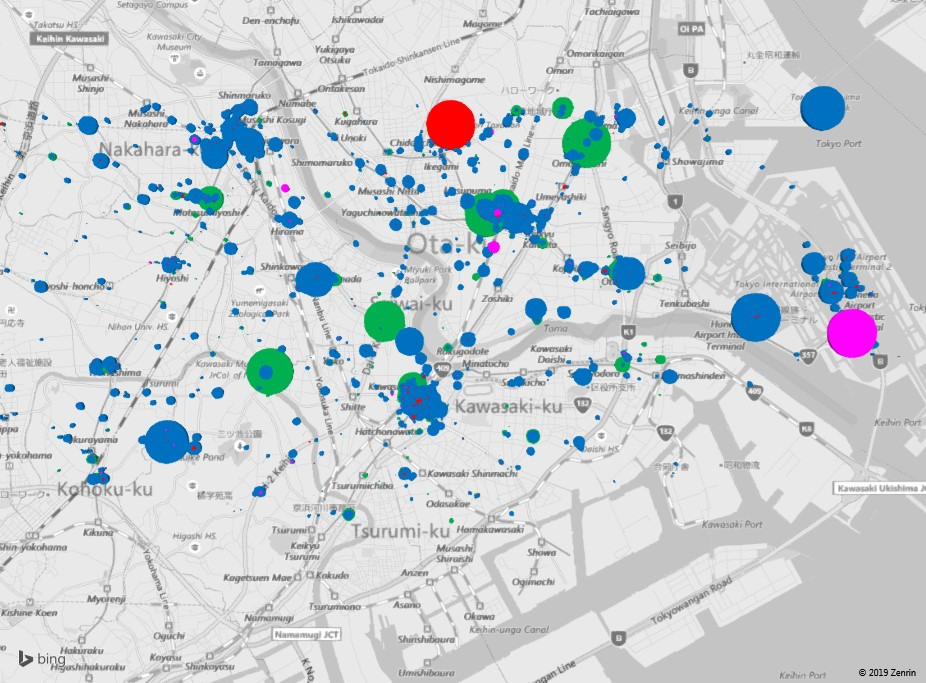}}}%
	\caption{Spatial patterns derived from the 2nd mode (location) of the tensor using Cut-CD-SC.}%
	\label{fig:spatial}
\end{figure*}

Using location factors $\boldsymbol{\mathrm{V}}$, the spatial patterns of D1 and D2 can be identified as shown in Fig.~\ref{fig:spatial}a and Fig.~\ref{fig:spatial}b respectively. An interesting pattern can be identified when the pink patterns from Fig.~\ref{fig:temporal}a and Fig.~\ref{fig:spatial}a are associated. The pattern can be interpreted as people who show similar kinds of activities over all 7 days, tend to be only active within their region. It should be noted that color sizes indicate the distribution within each pattern and does not necessarily show the overall dominance. For example, the green factor in Fig.~\ref{fig:spatial}a seems to be very active, but it means that the pattern is scattered. Connecting the color patterns of Fig.~\ref{fig:temporal}b and Fig.~\ref{fig:spatial}b, the red pattern shows 3 hot spots which are spatio-temporal patterns indicating that people tend to visit 3 places regularly around 8 am to 6 pm.

%\begin{figure*}[!t]
%\centering
%\begin{minipage}{.33\textwidth}
%  \centering
%  \includegraphics[width=5.2cm, height=4cm]{images/d1_temporal.png}
%  \captionof{figure}{Time factors on D1}
%  \label{fig:p1}
%\end{minipage}%
%\begin{minipage}{.33\textwidth}
%  \centering
%  \includegraphics[width=5.2cm, height=4cm]{images/d2_temporal.png}
%  \captionof{figure}{Time factors on D2}
%  \label{fig:p2}
%\end{minipage}
%\begin{minipage}{.33\textwidth}
%  \centering
%  \includegraphics[width=5.2cm]{images/d3_temporal.png}
%  \captionof{figure}{Time factors on D3}
%  \label{fig:p3}
%\end{minipage}
%\end{figure*}

%\begin{figure*}[!t]
%\centering
%\begin{minipage}{.33\textwidth}
%  \centering
%  \includegraphics[width=5.2cm, height=4cm]{images/d4_temporal.png}
%  \captionof{figure}{Time factors on D4}
%  \label{fig:p4}
%\end{minipage}%
%\begin{minipage}{.33\textwidth}
%  \centering
%  \includegraphics[width=4.7cm, height=3.3cm]{images/D1_spatial.png}
 % \includegraphics[width=5.2cm, height=4cm]{fig17.png}
%  \captionof{figure}{Spatial Pattern on D1}https://www.overleaf.com/3411893283gsqhhvfzjjng
%  \label{fig:p5}
%\end{minipage}
%\begin{minipage}{.33\textwidth}
%  \centering
%  \includegraphics[width=4.7cm, height=3.3cm]{images/D1_BPanjang.png}
%   % \includegraphics[width=5.2cm, height=4cm]{fig17.png}
%  \captionof{figure}{The Bukit Panjang Region}
%  \label{fig:p6}
%\end{minipage}
%\end{figure*}

%\begin{figure*}[!t]
%\centering
%\begin{minipage}{.33\textwidth}
%  \centering
%  \includegraphics[width=4.7cm, height=3.3cm]{images/D2_spatial.png}
%  \includegraphics[width=5.2cm, height=4cm]{fig19.png}
%  \captionof{figure}{Spatial Pattern on D2}
%  \label{fig:p7}
%\end{minipage}%
%\begin{minipage}{.33\textwidth}
%  \centering
%  \includegraphics[width=4.7cm, height=3.3cm]{images/D3_spatial.png}
   %\includegraphics[width=5.2cm, height=4cm]{fig20.png}
%  \captionof{figure}{Spatial Pattern on D3}
%  \label{fig:p8}
%\end{minipage}
%\begin{minipage}{.33\textwidth}
%  \centering
%  \includegraphics[width=4.7cm, height=3.3cm]{images/D4_spatial.png}
  %\includegraphics[width=5.2cm, height=4cm]{fig21.png}
%  \captionof{figure}{Spatial Pattern on D4}
%  \label{fig:p9}
%\end{minipage}
%\end{figure*}

%--------------
\subsubsection{Understanding the Spatio-temporal Patterns of Tokyo City Foursquare Users (D3 and D4)}

Fig.~\ref{fig:temporal}c and Fig.~\ref{fig:temporal}d show the temporal patterns of Cut-CD on D3 and D4 involving Tokyo city foursquare users. For D3 and D4, we set the rank to 2 and 4 respectively. We set the rank to 2, as there are no more unique patterns found in the dataset. Fig.~\ref{fig:temporal}c shows that Foursquare users are highly active on weekdays and moderately active on weekends. %This dataset is completely different from D1 and D2 which involve elderly people. This is reflected in Fig.~\ref{fig:temporal}c where we can identify only 2 patterns over seven days of the week. While elderly people have shown high activity on weekends, Foursquare users are highly active on weekdays and moderately active on weekends. 
Fig.~\ref{fig:spatial}c shows the respective spatial patterns associated with red and blue temporal patterns in Fig.~\ref{fig:temporal}c. While looking at Fig.~\ref{fig:temporal}d, the red pattern shows a peak between 8 $-$ 10 am and 6 $-$ 9 pm. The Pink pattern shows a peak between  8 $-$ 1 pm and 6 $-$ 12 am.  While these patterns occur only at few hot spots as shown in Fig.~\ref{fig:spatial}d, blue and green patterns showing a peak at 10 pm and 9 am are widespread.

\subsubsection{Pattern Distinctiveness} 
As shown in Table~\ref{table_pd}, Cut-CD-SC outperforms all the benchmarks by generating more unique and meaningful patterns. The column-wise element selection of Cut-CD-SC enables it to find unique patterns. ALS ran out of memory for D3 whereas both ALS and Turbo-SMT-OPT ran out of memory for D4. Cut-CD-SC does not show a significant improvement for D3. This is due to the presence of very few unique patterns in the dataset (Fig.~\ref{fig:temporal}c). For datasets D2 and D4 where there exist more patterns, Cut-CD-SC has significantly generated unique and meaningful patterns. 

\begin{table}[tb!]
\caption{Pattern distinctiveness (smaller is desirable).}
\label{table_pd}
\centering
\begin{tabular}{lccccc}
\hline
Method & D1 & D2 & D3 & D4 & Avg.\\
\hline
ALS   & 0.87 &  0.45  & o.m  &  o.m & - \\
Turbo-SMT OPT    & 0.82 &  0.49  &\textbf{ 0.98}  &  o.m & -\\
CCD++1	& 0.85	& 0.50	& 0.99	& 0.62 & 0.74\\
CCD++3	& 0.85	& 0.50	& 0.99	& 0.62 & 0.74\\
GCD     & 0.87 &  0.49  & 0.99  &  0.66 & 0.75\\
Cut-CD    & 0.64 &  0.49  & 0.99  & 0.59 & 0.67\\
Cut-CD-SC   & \textbf{0.63} &  \textbf{0.41}  & 0.99  &  \textbf{0.47} & \textbf{0.62}\\
\hline
\end{tabular}
\end{table}

\begin{figure}[ht!]
	\centering
	\subfloat[Cut-CD SC]{{\includegraphics[width=1.6in, height=1.34in]{images/tmp4.png}}}%
	\qquad	
	\subfloat[Cut-CD]{{\includegraphics[width=1.6in, height=1.34in]{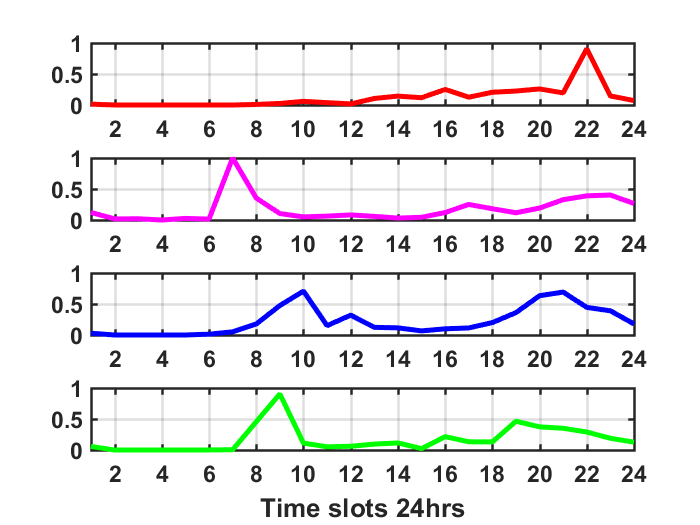}}}%
	\caption{Temporal patterns derived from the 3rd mode (time) of the tensor on D4 using Cut-CD-SC and Cut-CD showing the evidence of simultaneous elimination problem.}%
	\label{fig:comp}
\end{figure}

In Fig.~\ref{fig:comp}, we compare the patterns derived from Cut-CD and Cut-CD-SC for D4. The red, pink and blue patterns in Fig.~\ref{fig:comp}a can be paired up with green, blue and red patterns respectively in Fig.~\ref{fig:comp}b. %Similarly, the pink pattern in Fig.~\ref{fig:comp}a is similar to the blue pattern in Fig.~\ref{fig:comp}b while blue pattern in Fig.~\ref{fig:comp}a is similar to red pattern in Fig.~\ref{fig:comp}b showing a peak only at 10 pm.
Cut-CD in Fig.~\ref{fig:comp}b identifies 1 more pattern (i.e. pink) that is highly similar to blue pattern. This shows the inability of Cut-CD to avoid the simultaneous elimination problem (i.e., a state where similar patterns are derived multiple times)\cite{zou2008f}. On the other hand, Cut-CD-SC avoids generating the same patterns repeatedly due to the application of sparsity constraint and therefore, the green pattern in Fig.~\ref{fig:comp}a is a straight line stating there are no more unique patterns.

%==================
\section{Conclusion}
An element selection-based CD algorithm, Cut-CD, has been introduced to improve the efficiency of the N-CMTF computation. Cut-CD facilitates efficient factorization to reveal useful information in multifaceted datasets. Cut-CD first selects elements in a factor matrix according to the element importance based on the proposed column-wise cut-off technique. All the selected elements of the factor matrix are then updated. %As N-CMTF fails to generate true patterns in sparse datasets, a sparsity constrained N-CMTF called CAN-CSMTF has further been developed to derive spatio-temporal patterns. 
We conducted theoretical and empirical studies to demonstrate the effectiveness of Cut-CD. Theoretical analysis shows that the computational efficiency is achieved by Cut-CD by avoiding frequent gradient updates and proves the convergence property of Cut-CD. Empirical analysis shows that Cut-CD outperforms existing state-of-the-art algorithms in terms of scalability and convergence speed without compromising accuracy. It performs well, especially in sparse data conditions. Cut-CD has been shown to be efficient not only in missing value identification but also in the identification of factor problems with the incorporation of a sparsity constraint on factor matrices. In the future, we will explore the Cut-CD for the reduction of the computational complexity of MTTKRP.
% More importantly, Cut-CD is not restricted to N-CMTF model only, it can be easily  adapted for general NMF and NTF problems. 
%In future, we will extend the Cut-CD algorithm for partially shared matrix tensor factorization.

\section*{Acknowledgment}
We will like to thank the Lee Kuan Yew Centre for Innovative Cities (Aging Urbanism and NSFC 61750110529) to provide us the Singapore Elderly dataset.
\ifCLASSOPTIONcaptionsoff
  \newpage
\fi

\bibliographystyle{IEEEtran}
\bibliography{main}

\begin{thebibliography}{10}
\providecommand{\url}[1]{#1}
\csname url@samestyle\endcsname
\providecommand{\newblock}{\relax}
\providecommand{\bibinfo}[2]{#2}
\providecommand{\BIBentrySTDinterwordspacing}{\spaceskip=0pt\relax}
\providecommand{\BIBentryALTinterwordstretchfactor}{4}
\providecommand{\BIBentryALTinterwordspacing}{\spaceskip=\fontdimen2\font plus
\BIBentryALTinterwordstretchfactor\fontdimen3\font minus
  \fontdimen4\font\relax}
\providecommand{\BIBforeignlanguage}[2]{{%
\expandafter\ifx\csname l@#1\endcsname\relax
\typeout{** WARNING: IEEEtran.bst: No hyphenation pattern has been}%
\typeout{** loaded for the language `#1'. Using the pattern for}%
\typeout{** the default language instead.}%
\else
\language=\csname l@#1\endcsname
\fi
#2}}
\providecommand{\BIBdecl}{\relax}
\BIBdecl

\bibitem{acar2011all}
E.~Acar, T.~G. Kolda, and D.~M. Dunlavy, ``All-at-once optimization for coupled
  matrix and tensor factorizations,'' \emph{arXiv preprint arXiv:1105.3422},
  2011.

\bibitem{balasubramaniam2018nonnegative}
T.~Balasubramaniam, R.~Nayak, and C.~Yuen, ``Nonnegative coupled matrix tensor
  factorization for smart city spatiotemporal pattern mining,'' in
  \emph{Machine Learning, Optimization, and Data Science}.\hskip 1em plus 0.5em
  minus 0.4em\relax Springer, 2019, pp. 520--532.

\bibitem{acar2015data}
E.~Acar, R.~Bro, and A.~K. Smilde, ``Data fusion in metabolomics using coupled
  matrix and tensor factorizations,'' \emph{Proceedings of the IEEE}, vol. 103,
  no.~9, pp. 1602--1620, 2015.

\bibitem{acar2013understanding}
E.~Acar, M.~A. Rasmussen, F.~Savorani, T.~N{\ae}s, and R.~Bro, ``Understanding
  data fusion within the framework of coupled matrix and tensor
  factorizations,'' \emph{Chemometrics and Intelligent Laboratory Systems},
  vol. 129, pp. 53--63, 2013.

\bibitem{hunyadi2014block}
B.~Hunyadi, D.~Camps, L.~Sorber, W.~Van~Paesschen, M.~De~Vos, S.~Van~Huffel,
  and L.~De~Lathauwer, ``Block term decomposition for modelling epileptic
  seizures,'' \emph{EURASIP Journal on Advances in Signal Processing}, vol.
  2014, no.~1, p. 139, 2014.

\bibitem{cichocki2009nonnegative}
A.~Cichocki, R.~Zdunek, A.~H. Phan, and S.-i. Amari, \emph{Nonnegative matrix
  and tensor factorizations: applications to exploratory multi-way data
  analysis and blind source separation}.\hskip 1em plus 0.5em minus 0.4em\relax
  John Wiley \& Sons, 2009.

\bibitem{iakovidou2010multiway}
N.~Iakovidou, P.~Symeonidis, and Y.~Manolopoulos, ``Multiway spectral
  clustering link prediction in protein-protein interaction networks,'' in
  \emph{ITAB}.\hskip 1em plus 0.5em minus 0.4em\relax IEEE, 2010, pp. 1--4.

\bibitem{zinkevich2010parallelized}
M.~Zinkevich, M.~Weimer, L.~Li, and A.~J. Smola, ``Parallelized stochastic
  gradient descent,'' in \emph{NIPS}, 2010, pp. 2595--2603.

\bibitem{lee2001algorithms}
D.~D. Lee and H.~S. Seung, ``Algorithms for non-negative matrix
  factorization,'' in \emph{NIPS}, 2001, pp. 556--562.

\bibitem{wright2015coordinate}
S.~J. Wright, ``Coordinate descent algorithms,'' \emph{Mathematical
  Programming}, vol. 151, no.~1, pp. 3--34, 2015.

\bibitem{cai2011graph}
D.~Cai, X.~He, J.~Han, and T.~S. Huang, ``Graph regularized nonnegative matrix
  factorization for data representation,'' \emph{IEEE TPAMI}, vol.~33, no.~8,
  pp. 1548--1560, 2011.

\bibitem{turbo}
E.~E. Papalexakis, C.~Faloutsos, T.~M. Mitchell, P.~P. Talukdar, N.~D.
  Sidiropoulos, and B.~Murphy, ``Turbo-smt: Accelerating coupled sparse
  matrix-tensor factorizations by 200x,'' in \emph{SDM}.\hskip 1em plus 0.5em
  minus 0.4em\relax SIAM, 2014, pp. 118--126.

\bibitem{bader2007efficient}
B.~W. Bader and T.~G. Kolda, ``Efficient matlab computations with sparse and
  factored tensors,'' \emph{SIAM Journal on Scientific Computing}, vol.~30,
  no.~1, pp. 205--231, 2007.

\bibitem{flexifact}
A.~Beutel, P.~P. Talukdar, A.~Kumar, C.~Faloutsos, E.~E. Papalexakis, and E.~P.
  Xing, ``Flexifact: Scalable flexible factorization of coupled tensors on
  hadoop,'' in \emph{SDM}.\hskip 1em plus 0.5em minus 0.4em\relax SIAM, 2014,
  pp. 109--117.

\bibitem{papalexakis2013scoup}
E.~E. Papalexakis, T.~M. Mitchell, N.~D. Sidiropoulos, C.~Faloutsos, P.~P.
  Talukdar, and B.~Murphy, ``Scoup-smt: Scalable coupled sparse matrix-tensor
  factorization,'' \emph{arXiv preprint arXiv:1302.7043}, 2013.

\bibitem{oh2017s}
J.~Oh, K.~Shin, E.~E. Papalexakis, C.~Faloutsos, and H.~Yu, ``S-hot: Scalable
  high-order tucker decomposition,'' in \emph{WSDM}.\hskip 1em plus 0.5em minus
  0.4em\relax ACM, 2017, pp. 761--770.

\bibitem{shin2017fully}
K.~Shin, L.~Sael, and U.~Kang, ``Fully scalable methods for distributed tensor
  factorization,'' \emph{IEEE TKDE}, vol.~29, no.~1, pp. 100--113, 2017.

\bibitem{gcd2011}
C.-J. Hsieh and I.~S. Dhillon, ``Fast coordinate descent methods with variable
  selection for non-negative matrix factorization,'' in \emph{SIGKDD}.\hskip
  1em plus 0.5em minus 0.4em\relax ACM, 2011, pp. 1064--1072.

\bibitem{bhargava2015and}
P.~Bhargava, T.~Phan, J.~Zhou, and J.~Lee, ``Who, what, when, and where:
  Multi-dimensional collaborative recommendations using tensor factorization on
  sparse user-generated data,'' in \emph{International conference on world wide
  web}, 2015, pp. 130--140.

\bibitem{zheng2014diagnosing}
Y.~Zheng, T.~Liu, Y.~Wang, Y.~Zhu, Y.~Liu, and E.~Chang, ``Diagnosing new york
  city's noises with ubiquitous data,'' in \emph{UbiComp}.\hskip 1em plus 0.5em
  minus 0.4em\relax ACM, 2014, pp. 715--725.

\bibitem{zhu2016context}
X.~Zhu and R.~Hao, ``Context-aware location recommendations with tensor
  factorization,'' in \emph{ICCC}.\hskip 1em plus 0.5em minus 0.4em\relax IEEE,
  2016, pp. 1--6.

\bibitem{acar2012metabolomics}
E.~Acar, G.~Gozde, A.~R. Morten, R.~Daniela, O.~D. Lars, and B.~Rasmus,
  ``Coupled matrix factorization with sparse factors to identify potential
  biomarkers in metabolomics,'' \emph{IJKDB}, vol.~3, no.~3, pp. 22--43, 2012.

\bibitem{comon2009tensor}
P.~Comon, X.~Luciani, and A.~L. De~Almeida, ``Tensor decompositions,
  alternating least squares and other tales,'' \emph{Journal of Chemometrics: A
  Journal of the Chemometrics Society}, vol.~23, no. 7-8, pp. 393--405, 2009.

\bibitem{kolda2009tensor}
T.~G. Kolda and B.~W. Bader, ``Tensor decompositions and applications,''
  \emph{SIAM review}, vol.~51, no.~3, pp. 455--500, 2009.

\bibitem{TTB_Sparse}
B.~W. Bader and T.~G. Kolda, ``Efficient {MATLAB} computations with sparse and
  factored tensors,'' \emph{SIAM Journal on Scientific Computing}, vol.~30,
  no.~1, pp. 205--231, December 2007.

\bibitem{imtiaz2018distributed}
H.~Imtiaz and A.~D. Sarwate, ``Distributed differentially private algorithms
  for matrix and tensor factorization,'' \emph{JSTSP}, vol.~12, no.~6, pp.
  1449--1464, 2018.

\bibitem{cicho2009fhals}
A.~Cichocki and A.-H. Phan, ``Fast local algorithms for large scale nonnegative
  matrix and tensor factorizations,'' \emph{IEICE transactions on fundamentals
  of electronics, communications and computer sciences}, vol.~92, 2009.

\bibitem{richtarik2018parallel}
P.~Richtárik and M.~Takáč, ``Parallel coordinate descent methods for big
  data optimization,'' \emph{Mathematical Programming}, pp. 433--484, 2016.

\bibitem{kim2014alg}
J.~Kim, Y.~He, and H.~Park, ``Algorithms for nonnegative matrix and tensor
  factorizations: a unified view based on block coordinate descent framework,''
  \emph{Journal of Global Optimization}, vol.~58, no.~2, pp. 285--319, 2014.

\bibitem{kimura2015var}
K.~Kimura and M.~Kudo, ``Variable selection for efficient nonnegative tensor
  factorization,'' in \emph{ICDM}.\hskip 1em plus 0.5em minus 0.4em\relax IEEE,
  2015, pp. 805--819.

\bibitem{ccdpp}
H.-F. Yu, C.-J. Hsieh, S.~Si, and I.~Dhillon, ``Scalable coordinate descent
  approaches to parallel matrix factorization for recommender systems,'' in
  \emph{ICDM}.\hskip 1em plus 0.5em minus 0.4em\relax IEEE, 2012, pp. 765--774.

\bibitem{shin2014distributed}
K.~Shin and U.~Kang, ``Distributed methods for high-dimensional and large-scale
  tensor factorization,'' in \emph{ICDM}.\hskip 1em plus 0.5em minus
  0.4em\relax IEEE, 2014, pp. 989--994.

\bibitem{rossi2016parallel}
R.~A. Rossi and R.~Zhou, ``Parallel collective factorization for modeling large
  heterogeneous networks,'' \emph{Social Network Analysis and Mining}, vol.~6,
  no.~1, p.~67, 2016.

\bibitem{chen2018adaptive}
M.~Chen, Q.~Wang, and X.~Li, ``Adaptive projected matrix factorization method
  for data clustering,'' \emph{Neurocomputing}, vol. 306, pp. 182--188, 2018.

\bibitem{luong2018novel}
K.~Luong, T.~Balasubramaniam, and R.~Nayak, ``A novel technique of using
  coupled matrix and greedy coordinate descent for multi-view data
  representation,'' in \emph{International Conference on Web Information
  Systems Engineering (WISE)}.\hskip 1em plus 0.5em minus 0.4em\relax Springer,
  2018, pp. 285--300.

\bibitem{luong2019clustering}
K.~Luong and R.~Nayak, ``Clustering multi-view data using non-negative matrix
  factorization and manifold learning for effective understanding: A survey
  paper,'' in \emph{Linking and Mining Heterogeneous and Multi-view
  Data}.\hskip 1em plus 0.5em minus 0.4em\relax Springer, 2019, pp. 201--227.

\bibitem{tucker1966}
Tucker and L.~R, ``Some mathematical notes on three-mode factor analysis,''
  \emph{Psychometrika}, vol.~31, no.~3, 1966.

\bibitem{carroll1970}
J.~D. Carroll and J.-J. Chang, ``Analysis of individual differences in
  multidimensional scaling via an n-way generalization of “eckart-young”
  decomposition,'' \emph{Psychometrika}, vol.~35, no.~3, pp. 283--319, 1970.

\bibitem{boyd2011distributed}
S.~Boyd, N.~Parikh, E.~Chu, B.~Peleato, J.~Eckstein \emph{et~al.},
  ``Distributed optimization and statistical learning via the alternating
  direction method of multipliers,'' \emph{Foundations and
  Trends{\textregistered} in Machine learning}, vol.~3, no.~1, pp. 1--122,
  2011.

\bibitem{jin2017escape}
C.~Jin, R.~Ge, P.~Netrapalli, S.~M. Kakade, and M.~I. Jordan, ``How to escape
  saddle points efficiently,'' in \emph{ICML}.\hskip 1em plus 0.5em minus
  0.4em\relax JMLR. org, 2017, pp. 1724--1732.

\bibitem{billy2017extract}
B.~P.~L. Lau, S.~H. Marakkalage, S.~K. Viswanath, T.~Balasubramaniam, Y.~Chau,
  Y.~Belinda, and R.~Nayak, ``Extracting point of interest and classifying
  environment for low sampling crowd sensing smartphone sensor data,'' in
  \emph{PerCom Workshops}.\hskip 1em plus 0.5em minus 0.4em\relax IEEE, 2017,
  pp. 201--206.

\bibitem{balasubramaniam2018understanding}
T.~Balasubramaniam, R.~Nayak, and C.~Yuen, ``Understanding urban
  spatio-temporal usage patterns using matrix tensor factorization,'' in
  \emph{ICDMW}.\hskip 1em plus 0.5em minus 0.4em\relax IEEE, 2018, pp.
  1497--1498.

\bibitem{zou2008f}
H.~Zou and M.~Yuan, ``The f$\infty$-norm support vector machine,''
  \emph{Statistica Sinica}, pp. 379--398, 2008.

\end{thebibliography}
\vskip -2.8\baselineskip plus -1fil
\begin{IEEEbiography}[{\includegraphics[width=1in,height=1.25in,clip,keepaspectratio]{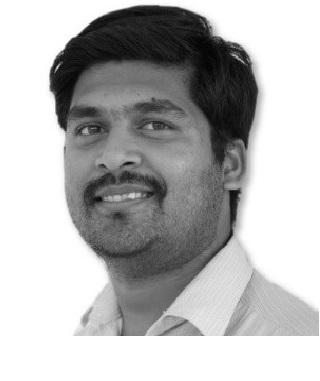}}]{Thirunavukarasu Balasubramaniam} received the B.E. degree from Anna University, Chennai, India, in 2015. He is currently pursuing the Ph.D. degree with the Queensland University of Technology (QUT), Brisbane, Australia. From 2015 to 2016, he was a Research Assistant with the SUTD, Singapore. His current research interests include machine learning, factorization, recommender systems, and pattern mining.
\end{IEEEbiography}
\vskip -3\baselineskip plus -1fil
\begin{IEEEbiography}
[{\includegraphics[width=1in,height=1.25in,clip,keepaspectratio]{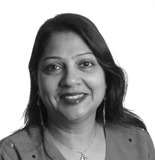}}]{Richi Nayak} received M.E. degree from the Indian Institute of Technology, Roorkee India in 1995 and PhD in Computer Science from QUT, Brisbane, Australia in 2001. She is currently Associate Professor of computer science at QUT. Her current research interests include machine learning, text mining, personalization, automation, and social network analysis.
\end{IEEEbiography}
\vskip -3.4\baselineskip plus -1fil
\begin{IEEEbiography}
[{\includegraphics[width=1in,height=1.25in,clip,keepaspectratio]{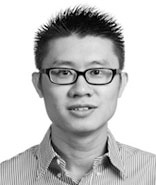}}]{Chau Yuen} received the B.Eng. and Ph.D. degrees from NTU, Singapore, in 2000 and 2004, respectively. He was a Post-Doc Fellow with the Lucent Technologies Bell Labs, NJ, USA, in 2005. He was a Visiting Assistant Professor with Hong Kong PolyU in 2008. From 2006 to 2010, he was a Senior Research Engineer with I2R, Singapore. He is currently an Associate Professor with the SUTD, Singapore. He is an Editor of the IEEE Trans. Commun. and the IEEE Trans. Veh. Technol.
\end{IEEEbiography}
\vskip -2.3\baselineskip plus -1fil
\begin{IEEEbiography}
[{\includegraphics[width=1in,height=1.25in,clip,keepaspectratio]{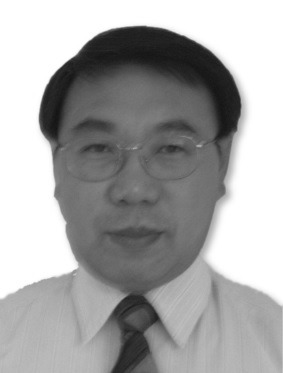}}]{Yu-Chu Tian} (M'00, SM'19) received the Ph.D. degree in computer and software engineering in 2009 from the University of Sydney, Australia, and the Ph.D. degree in industrial automation in 1993 from Zhejiang University, Hangzhou, China. He is currently a professor of computer science with QUT. His research interests include distributed computing, cloud computing, real-time computing and control systems.
\end{IEEEbiography}
\end{document}